\documentclass[10pt,twocolumn,letterpaper]{article}
\pdfoutput=1
\usepackage{confstyle}             

\usepackage{amsmath}    
\usepackage{algorithm}  
\usepackage{algpseudocode} 
\usepackage{amsfonts}   
\usepackage{multirow}
\usepackage{graphicx}
\usepackage{makecell}
\usepackage{bibunits}

\usepackage[utf8]{inputenc}
\usepackage{rotating}
\usepackage{pifont}
\usepackage{amsmath}    
\usepackage{algorithm}  
\usepackage{algpseudocode} 
\usepackage{amsfonts}
\usepackage{lineno}
\definecolor{arxivblue}{rgb}{0.21,0.49,0.74}
\usepackage[pagebackref,breaklinks,colorlinks,allcolors=arxivblue]{hyperref}

\title{
DLTPose: 6DoF Pose Estimation From Accurate Dense Surface Point Estimates}

\author{Akash Jadhav, Michael Greenspan\\
Dept. of Electrical and Computer Engineering, Ingenuity Labs Research Institute \\
Queen’s University, Kingston, Ontario, Canada}

\makeatletter
\providecommand{\@LN@col}[1]{}
\makeatother

\begin{document}
\maketitle

\begin{abstract}
We propose DLTPose, a novel method for 6DoF object pose estimation from RGBD images that combines the accuracy of sparse keypoint methods with the robustness of dense pixel-wise predictions. DLTPose predicts per-pixel radial distances to a set of minimally four keypoints, which are then fed into our novel Direct Linear Transform (DLT) formulation to produce accurate 3D object frame surface estimates, leading to better 6DoF pose estimation. Additionally, we introduce a novel symmetry-aware keypoint ordering approach, designed to handle object symmetries that otherwise cause inconsistencies in keypoint assignments. Previous keypoint-based methods relied on fixed keypoint orderings, which failed to account for the multiple valid configurations exhibited by symmetric objects, which our ordering approach exploits to enhance the model's ability to learn stable keypoint representations. Extensive experiments on the benchmark BOP-Classic-Core datasets show that DLTPose outperforms many recent methods, performing especially well for symmetric and occluded objects. The code is available at anonymous.4open.science/r/DLTPose
\end{abstract}
    
\section{Introduction}
\label{sec:intro}

Object pose estimation is a fundamental problem in computer vision with broad applications in robotics, augmented reality, and autonomous systems~\cite{Peng, Tremblay, Pavlakos}. The goal is to determine an object's six-degree-of-freedom (6DoF) pose, which comprises both a 3D rotation and a 3D translation, from visual data. This task is particularly challenging due to factors such as occlusions, background clutter, sensor noise, varying lighting conditions, and object symmetries, all of which introduce ambiguities and uncertainties.

\begin{figure}[t]
    \centering
    \begin{subfigure}{0.9\linewidth}
    \centering
    \includegraphics[height=4cm, width=0.9\linewidth]{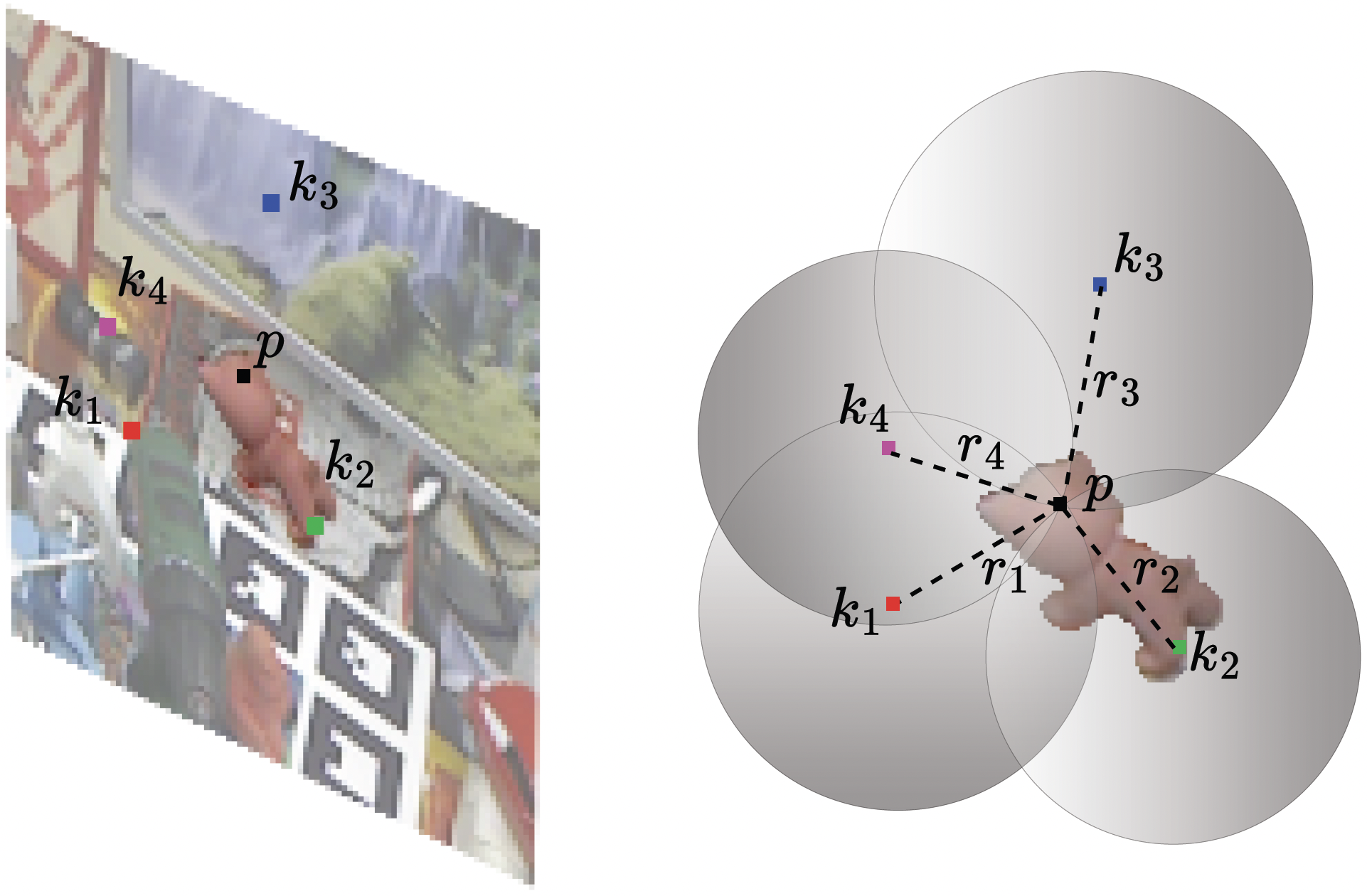} 
    \caption{}
    \label{fig:intro_1}
\end{subfigure}

\vspace{0.1cm} 

\begin{subfigure}{0.9\linewidth}
    \centering
    \includegraphics[height=2.7cm, width=0.68\linewidth]{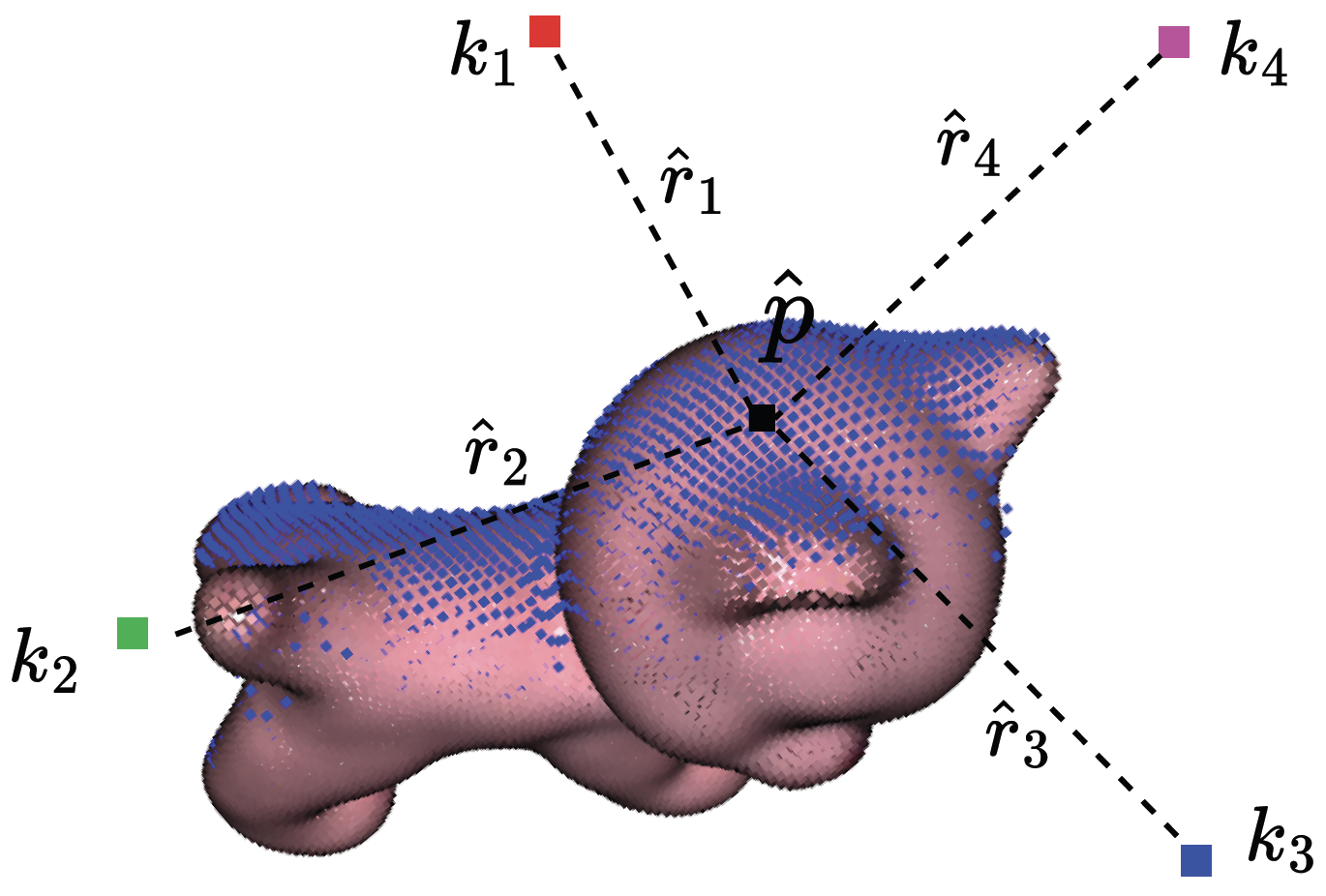} 
    \caption{}
    \label{fig:intro_2}
\end{subfigure}
    
    \caption{Visualization of DLTPose surface estimation. (a) For image point $p_i$, four radial distances $\hat{r}_1, \hat{r}_2, \hat{r}_3, \hat{r}_4$ are estimated per-pixel, as the Euclidean distance to four predefined keypoints $k_j$. The DLT solution uses these keypoints and radial distances to estimate object frame 3D surface points $\bar{p}_i$. (b)  Estimated surface points (blue) overlaid on the object mesh, in the object frame.}
\label{fig:intro}
\end{figure}

Among modern approaches, two dominant paradigms have emerged: sparse and dense methods~\cite{s24041076}. Sparse methods ~\cite{Wu, He, Peng} focus on predicting a small set of keypoints with high accuracy, which ultimately leads to increased accuracy of estimated poses. 
While they can be highly accurate, a limitation of sparse methods is that their reliance on a limited number of keypoints makes them more susceptible to occlusions, as missing or misidentified keypoints can significantly impact overall estimation. In contrast, dense methods ~\cite{Brachmann, Park, Haugaard, labbe2020} predict per-pixel 3D surface coordinates in the object frame using 2D image inputs, providing high robustness through redundant predictions. This redundancy helps mitigate errors by utilizing techniques such as RANSAC to filter out inconsistent and inaccurate predictions. While the increased number of surface points estimated can increase accuracy in occluded and cluttered scenes, dense methods have not specifically focused on obtaining highly accurate estimates of the 3D visible surface of the object. Instead, their primary objective has been to generate a broad and redundant representation, often at the expense of accurately localizing individual points.

This work proposes a unified approach that combines the accuracy of sparse with the redundancy of dense methods. 
Our method, called DLTPose, trains a CNN on RGB-D data to estimate a per-pixel (minimally) four-dimensional representation, where each channel encodes a radial distance, defined as the Euclidean distance between a 
3D scene point
corresponding to a 2D image pixel,  relative to a predefined 3D keypoint. A minimum of four keypoints is required to 
estimate the pixel surface coordinates in the object-centric reference frame, by solving a novel and highly accurate Direct Linear Transform (DLT) formulation.
In this way, the radial quantities inferred from the network are combined with the DLT method to produce an accurate 2D-3D estimate of the visible surface of the object.
The estimated object frame points, along with their corresponding 3D camera frame points, are then passed to a RANSAC-enabled Umeyama algorithm~\cite{umeyama} to estimate the final object pose.

This pipeline enables precise and robust dense 3D surface estimation, 
leading to improved pose accuracy compared to prior methods where 
dense 2D-3D surface points are inferred using end-to-end networks, albeit with lower accuracy~\cite{Park,labbe2020,Haugaard}.
By integrating the precision of sparse keypoint-based methods with the redundancy of dense approaches, the proposed framework enhances both robustness and accuracy of pose estimation. This unified strategy improves the fidelity of 3D surface reconstruction, which in turn leads to more reliable pose estimation, particularly in occluded scenes where conventional methods struggle.

Handling objects with inherent symmetries 
is challenging, as visually identical orientations can mislead network training and prediction. Standard loss functions often fail to address these symmetries, resulting in large errors when equivalent poses are treated as distinct. Previous methods attempt to mitigate this by restricting training poses~\cite{kehl, Markus} or mapping predictions to the nearest symmetric equivalent~\cite{rad2017bb8}, but these approaches struggle with discrete symmetries, leading to reduced generalization and inconsistencies. 
Rather than
limiting training diversity or applying post-hoc corrections, our approach inherently incorporates symmetry awareness by ensuring keypoints remain stable across symmetrical transformations. Unlike traditional keypoint-based methods, which assign fixed keypoints and suffer from inconsistencies under symmetric rotations, our method maintains stable keypoint relationships, improving pose accuracy and robustness across different viewpoints.

The key contributions of this paper are as follows: 
\begin{itemize}
\item 
We propose a novel DLT formulation to estimate accurate 3D object frame surface points from 2D image points by leveraging per-pixel radial distance predictions to a minimal set of four keypoints. The resulting accurate 3D surface estimates improve the accuracy of pose estimation, especially in difficult cluttered and occluded scenarios.
\item 
We introduce a symmetry-aware keypoint framework that dynamically reorders keypoints based on their relative position in the camera view, ensuring stable radial map learning across symmetric transformations. This prevents inconsistencies in keypoint assignments, reducing regression errors and improving robustness in pose estimation for symmetric objects.

\item 
Our method achieves state-of-the-art results compared to recent leading methods on benchmark datasets, demonstrating superior performance in handling occlusions, cluttered environments, and symmetric objects. 
\end{itemize}
\section{Literature Review}
\label{sec:review}

Deep learning-based pose estimation techniques
can be broadly classified into sparse and dense strategies~\cite{s24041076}, each with distinct advantages and challenges.
Sparse methods rely on detecting a small set of keypoints in an image and establishing correspondences with known 3D points defined in the object frame. 
BB8~\cite{rad2017bb8} was 
an early learning-based sparse method 
that
regressed 2D projections of objects' 3D bounding box corners, and then applied a PnP solver for pose estimation. 
KeyPose~\cite{sundermeyer2020augmented} introduced a structured approach to keypoint regression, leveraging data augmentation techniques to improve robustness. It extended the idea of bounding box-based keypoint detection by incorporating keypoint refinement strategies, making it more robust to perspective distortions. 
PVNet~\cite{Peng} 
regressed vector fields pointing toward keypoint locations, relying on a RANSAC-based voting strategy to filter out noisy predictions. Building on PVNet, PVN3D~\cite{He} extended keypoint regression to 3D by voting on clusters of offsets to keypoints within point cloud data. 
RCVPose~\cite{Wu}
extended the sparse approach by
accumulating votes on the surface of intersecting
spheres defined by radial values regressed for each image pixel. It was shown that the 1D radial value
resulted in more accurately localized keypoints compared against the 2D vector value of PVNet and the 3D offset value of PVN3D. 
Despite these advances, most sparse methods relied on heuristically chosen keypoints that may not generalize well across different object shapes. KeyGNet~\cite{wu2024keygnet} addressed this limitation by learning optimal keypoint locations. However, sparse methods remain susceptible to failures when keypoints are missing or misidentified due to occlusions or viewpoint changes.

Dense methods addressed these limitations by estimating per-pixel 3D surface points of objects, providing richer geometric information for pose estimation. 
In the early seminal work of~\cite{Brachman1},
a Random Forest was trained to estimate
dense 3D surface coordinates in the object reference
frame,
which in later work was refined with an energy minimization approach
using a DNN~\cite{Brachmann}.
In the Normalized Object Coordinate Space (NOCS) framework~\cite{wang2019normalized}, a canonical representation was proposed  to estimate dense surface coordinates by normalizing object shape and scale, 
thereby improving learning generalization across different objects. 
Pix2Pose~\cite{Park} built upon this concept by regressing per-pixel object coordinate predictions
through a 
fully convolutional backbone,
refining the resulting
RANSAC PnP derived pose estimates through the use of confidence maps.

Building upon the dense approach, 
PoseCNN~\cite{xiang2018posecnn} directly regresses the object center, depth, and quaternion for each detected region-of-interest.
Coupled Iterative Refinement~\cite{lipson2022coupled} introduced a multi-stage process that refined pose predictions by progressively aligning estimated 3D coordinates with observed depth data, mitigating initial pose estimation errors and enhancing robustness in challenging scenes. CosyPose~\cite{labbe2020} employs iterative render-and-compare refinement with continuous rotation parameterization, explicit symmetry handling, and modern backbones for improved scalability. 
ZebraPose~\cite{su2022zebrapose} learns region-specific binary codes for object vertices, with vertices recursively partitioned and assigned descriptors. A per-object network predicts the binary code of the corresponding 3D vertex for each pixel, yielding dense 2D–3D correspondences for pose estimation. 
IRPE~\cite{Jin2024IRPEIR} introduces an instance-level reconstruction based framework that leverages self-supervised learning to encode high-level object features, which are then used in a multi-task scheme for direct 6D pose regression, improving robustness under occlusion and clutter.
SurfEmb~\cite{Haugaard} introduced a surface embedding-based approach that learned a continuous representation of object surfaces instead of discrete coordinate mappings. This method improved surface correspondence estimation by embedding geometric features in a high-dimensional space, allowing for more accurate object localization, even in occluded or textureless regions.

Hybrid approaches attempt to integrate the benefits of both sparse and dense methods by leveraging the accuracy of keypoints while incorporating dense feature representations. FFB6D~\cite{he2021ffb6d} fuses feature-based keypoint detection with dense depth representations to improve robustness in cluttered and occluded environments. 
HiPose~\cite{lin2024hipose} introduced a hierarchical binary surface encoding that extracts dense RGB-D features and applies correspondence pruning, resulting in a reliable yet compact set of 3D-3D correspondences for pose estimation. By progressively filtering spurious matches through its hierarchical pruning strategy, HiPose achieves more accurate and efficient correspondence selection compared to purely dense regression methods.
DFTr~\cite{zhou2023deep} uses a Deep Fusion Transformer that models long-range RGB–depth correlations and introduces a weighted vector-wise voting scheme for more precise 3D keypoint localization.
Both HiPose and DFTr are build upon the fusion paradigm of FFB6D, extending it respectively with hierarchical correspondence pruning and transformer-based cross-modal modeling.

Our method leverages elements of both sparse and dense methods by integrating per-object pixel 3D surface estimates
from a set of (minimally four) keypoints, 
using a 
novel Direct Linear Transform (DLT) formulation.
Our framework is similar to that of Pix2Pose,
with our highly accurate DLT surface estimation
and our treatment of object symmetries
leading to improved pose estimation accuracy.
\section{Method}

\subsection{DLT Surface Estimation}

In keypoint methods, 
quantities regressed for
each foreground image pixel
are aggregated to estimate
a sparse set of 
keypoints in the camera frame.
For example, in RCVPose~\cite{Wu},
the radial distances from each pixel to each keypoint are inferred, and are combined with the pixels' corresponding depth values to vote upon the keypoints' 3D coordinates in the scene frame.
In this work, we propose an inverse process, 
whereby the radial quantities inferred at each image pixel
are combined with known object frame keypoint coordinates,
to estimate the corresponding object frame 3D pixel coordinates. In this way, the same inferred radial values that were previously used to estimate image frame keypoints~\cite{Wu}, are repurposed here for 3D surface reconstruction in the object frame.

Specifically,
let 
$\overline{\cal{O}}$
be an object
defined within its own object coordinate reference frame,
with
3D keypoint 
$\overline{k}_j$
defined within this same frame.
$\overline{\cal{O}}$
is transformed
by pose 
$[\boldsymbol{\mathcal{R}}|\textbf{t}]$
to reside within the image frame,
i.e. ${\cal{O}}\!=\!\boldsymbol{\mathcal{R}}
\ \overline{\cal{O}}+\textbf{t}$.
Let ${p}$ be the image frame point
corresponding to
$\overline{p}$
on the surface of $\overline{\cal{O}}$,
and let 
${k}_j$
be the image frame coordinate of object frame keypoint $\overline{k}_j$,
i.e. 
${p}\!=\!\boldsymbol{\mathcal{R}}
\ \overline{p}+\textbf{t}$
and
${k}_j\!=\!\boldsymbol{\mathcal{R}}
\ \overline{k}_j+\textbf{t}$.

At inference, a network estimates the radial (i.e. Euclidean) distance
$\widehat{r}_{j}$ between each $p$ and $k_j$. 
As rigid transformations are isometric and preserve distances,
this radial value is therefore also an estimate of the
distance between corresponding object frame surface points and keypoints:
\begin{equation}
\widehat{r}_{j} \simeq r_{j} = |\!|p-k_j|\!| 
= |\!|\overline{p}-\overline{k}_j|\!|.
\label{eq:r_ij}
\end{equation}

Expanding Eq.~\ref{eq:r_ij} and collecting terms (as derived in Sec.~\ref{sec:DLT Derivation} of the Supplementary Material) gives:
\newcommand{\s}{\hspace{0pt}}
\newcommand{\rs}{\!\!\!\!\!\!\!\!}
\newcommand{\rss}{\!\!\!\!}
\newcommand{\rsss}{\!\!}
\newcommand{\mdots}{\hspace{-10 pt} \cdots \hspace{-10 pt}}
\begin{equation}
\label{eq:matrix3}
\begin{bmatrix}
\!\! -2\overline{x}_{k_1} \rs 
& \rss -2\overline{y}_{k_1} \rs
& \rs -2\overline{z}_{k_1} \rs
& \rss 1 
& \rss
(|\!|\overline{k}_1|\!|^2 \!-\! \widehat{r}_{1}^2) \\
\vdots & \vdots & \vdots & \rss \vdots & \vdots \\
\! -2\overline{x}_{k_{N_k}} \rs
& -2\overline{y}_{k_{N_k}} \rs
& -2\overline{z}_{k_{N_k}} \rss
& \rss 1 
& \rss
(|\!|\overline{k}_{N_k}|\!|^2 \!-\! \widehat{r}_{N_k}^2) \\
\end{bmatrix}
\begin{bmatrix}
\overline{x} \\
\overline{y} \\
\overline{z} \\
|\!|\overline{p}|\!|^2 \\
1
\end{bmatrix}
= 0.
\end{equation}
Eq.~\ref{eq:matrix3} is in the familiar $\mathbf{A\overline{X}}\!\!=\!\!0$ form of the Direct Linear Transform (DLT)~\cite{Hartley:2003:MVG:861369}.
Matrix $\mathbf{A}$  comprises the known coordinates of
$N_k$
object frame keypoints 
$\{\overline{k}_j\}_{j=1}^{N_k}$
and their corresponding radial values $\hat{r}_{j}$ inferred from image frame point $p$, 
whereas $\mathbf{\overline{X}}$ comprises the unknown 
object frame coordinates $\overline{p}$ of $p$.
Composing $\mathbf{A}$ from $N_k \ge 4$ 
non-coplanar keypoints and their respective radial estimates
and performing Singular Value Decomposition,
the resulting Eigenvector corresponding to the smallest Eigenvalue yields
a least square estimate $\mathbf{\widehat{\overline{X}}}$ of 
$\mathbf{\overline{X}}$ 
up to an unknown scale,
which is recovered as
 the final row of 
 $\mathbf{\widehat{\overline{X}}}$.
 Eq.~\ref{eq:matrix3} is equivalent to estimating the 
 point 
 $\widehat{\overline{p}}$ of 
 $N_k \ge 4$ 
 intersecting spheres
 centered respectively at 
 $\overline{k}_j$
 with radii
 $\hat{r}_{j}$,
 as in Fig.~\ref{fig:intro}(a).
 A Hough solution to this sphere 
 intersection problem was first proposed
 in \cite{Wu}, 
 albeit in the discrete domain and therefore less accurate.
 Eq.~\ref{eq:matrix3} can be solved for all points
 $p$ that lie within an object's visibility mask output from segmentation. 
 This
 results in a 3D estimate of the visible surface of the object, expressed within the object frame, as shown in Fig.~\ref{fig:intro}(b).

 \subsection{Network Architecture}

\begin{figure*}[t]
    \centering
    \includegraphics[height=9.5cm,width=0.75\linewidth]{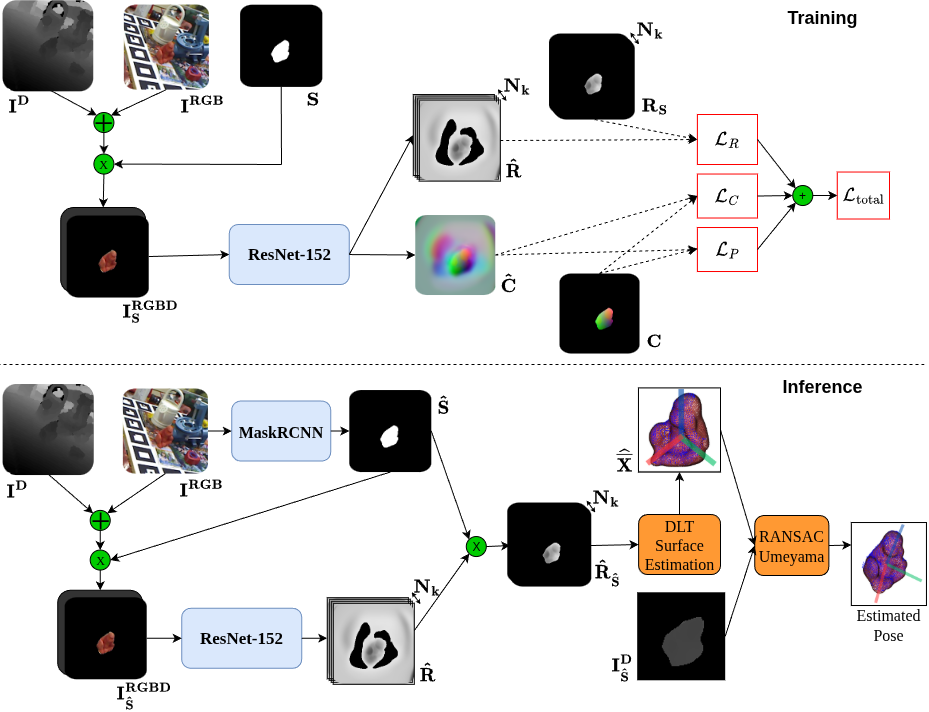}
    \caption{DLT network architecture. Both training and inference estimate radial map $\hat{R}$, which has channel depth $N_k \ge 4$ corresponding to the number of keypoints. At least 4 keypoints are required to solve the DLT formulation (Eq.~\ref{eq:matrix3}) to estimate object surface points $\mathbf{\widehat{\overline{X}}}$.}
    \label{fig:method}
\end{figure*}

In the network architecture illustrated in Fig.~\ref{fig:method}, the input 
at training consists of: 
the segmented object in the RGBD image 
$\mathbf{I}_S^{RGBD}$; 
$N_k \ge 4$ ground truth keypoint coordinates $k_{\text{j}}$
and their corresponding
segmented radial maps
$\mathbf{R}_{S}$,
and; the ground truth normalized point cloud image $\mathbf{C}$ for the object in its current pose.
For each object, 
the keypoints 
are either generated by
KeyGNet~\cite{wu2024keygnet}, 
or the proposed symmetric keypoints, as described in Sec.~\ref{sec:Symmetric Keypoints}. 
The ground truth radial map data 
$\mathbf{R}_S$ is a 
$H\!\!\times\!\!W\!\!\times\!\!N_k$ tensor comprising the radial distances between the 3D scene point corresponding to each visible 2D image pixel, relative to the $N_k$ predefined 3D keypoints,
where the 2D slice 
$\mathbf{R}_S^j = \mathbf{R}_S[:,:,j]$ 
represents 
the  $j^{th}$ keypoint.
The network's output consists of the 
unsegmented estimate 
{$\mathbf{\hat{R}}$} of
$\mathbf{R}_S$. 

The normalized object point cloud $\mathbf{C}$ is defined for an object's canonical pose, with its values scaled to the range $[0,1]$, following the 
approach
of
NOCS~\cite{wang2019normalized} 
and 
Pix2Pose~\cite{Park}.
When the model regresses $\mathbf{\hat{C}}$ to approximate $\mathbf{C}$, it provides an estimate of the object's shape in the canonical space. Although not perfectly accurate due to regression errors and projection ambiguities, it preserves essential geometric structures. This normalization aids in interpreting object geometry, 
and contributes to learning more accurate radial maps.

The network structure employs a ResNet-152 backbone~\cite{He_}, similar to PVNet~\cite{Peng}, with two key differences. First, we replaced LeakyReLU with ReLU as the activation function because our radial voting scheme only includes positive values, unlike the vector voting scheme of PVNet which also required accommodating negative values. Second, we increased the number of skip connections linking the downsampling and upsampling layers from three to five, allowing for the inclusion of a richer collection of additional local features during upsampling~\cite{Long}. 

The loss includes radial map regression term $\mathcal{L}_{R}$, which is the mean absolute error of the estimated radial values:
\begin{equation}
    \mathcal{L}_{R} = \frac{1}{N} 
    \sum_{p_i \in \mathbf{S}} 
    \left( \left| \hat{r}_{i} - r_{i} \right| \right)
    \label{eq:l_r}
\end{equation}
with 
the summation taking place 
for radial values
corresponding to all
$N$ points $p_i$ in segmentation mask $\mathbf{S}$.
There is also a soft $L1$ regression loss term 
$\mathcal{L}_{C}$
for the normalized object point cloud in the current pose, as proposed in~\cite{wang2019normalized}:
\begin{equation}
\mathcal{L}_{C} = \frac{1}{N} 
  \sum_{p_i \in \mathbf{S}}
\left\{
\begin{array}{ll}
5 \cdot (c_i - \hat{c_i})^2 & \text{if } |c_i - \hat{c_i}| \leq 0.1 \\
|c_i - \hat{c_i}| - 0.05 & \text{if } |c_i - \hat{c_i}| > 0.1
\end{array}
\right.
\label{eq:l_c}
\end{equation}
We
also include a
variation of the symmetric  loss
proposed in~\cite{wang2019normalized}.
This loss minimizes the difference between 
the
normalized coordinate map
$\mathbf{C}$
with its estimated value
$\hat{\mathbf{C}}_S$,
for one pose in the pose symmetry set $\Theta$:
\begin{equation}
    \mathcal{L}_{P} = \min_{\Theta} 
    (
    \frac{1}{N} 
      \sum_{p_i \in \mathbf{S}}
    \left(  \lfloor{\hat{c_i} \cdot n_{\text{b}}}\rfloor  -  \lfloor{c_i} \cdot n_{\text{b}}\rfloor \right)^2 
    )
    \label{eq:l_sc}
\end{equation}

Unlike~\cite{wang2019normalized},
we discretize the 
normalized coordinate values
to fall within a coarse discrete
range $[0,\ldots,n_\text{b}-1]$
for each of the three dimensions of normalized space.
We found in practise that this discretization served to improve performance for nearly symmetric objects 
(such as the LINEMOD ``glue'' object), but not for truly symmetric objects 
(such as ``eggbox'')
which suffered from the regular ambiguities,
as demonstrated in the ablation study in the Supplementary Material 
(Sec.~\ref{sec:ablation_experiments}, Tab.~\ref{tab:diff_loss_performance}).
For this reason, we call 
$\mathcal{L}_{P}$
the \emph{pseudo-symmetric} loss.
This loss term ensures that nearly symmetric objects are correctly aligned under the set 
$\Theta$
of pseudo-symmetric rotations, which improves the accuracy of radial regression.
The total loss \( \mathcal{L}_{\text{total}} \) is calculated as a weighted sum of the three distinct components:
\begin{equation}
\mathcal{L}_{\text{total}} = \lambda_1 \cdot \mathcal{L}_R + \lambda_2 \cdot \mathcal{L}_{C} + \lambda_3 \cdot \mathcal{L}_{P} 
    \label{eq:t_loss}
\end{equation}
For all experiments, the
scale values were 
set empirically to $\lambda_1\!=\!0.6$,  $\lambda_2\!=\!0.2 $ and  $\lambda_3\!=\!0.2$ for asymmetric and pseudo-symmetric objects. For symmetric objects, $\lambda_2$ and $\lambda_3$ were set to zero. 
The relative impact of the three loss terms 
is shown in the 
Supplementary Material, (Sec.~\ref{sec:ablation_experiments}).

During inference, 
the RGB image $\mathbf{I}^{RGB}$ is first processed by MaskRCNN~\cite{maskrcnn} to generate the segmented object mask $\hat{\mathbf{S}}$, 
as shown in Fig.~\ref{fig:method}.
$\hat{\mathbf{S}}$ is then element-wise multiplied by
$\mathbf{I}^{RGB}$
and depth image
$\mathbf{I}^{D}$
to produce the segmented RGBD object image $\mathbf{I}_{\hat{S}}^{RGBD}$. This segmented image is fed into ResNet-152 to produce the radial map estimates $\hat{\mathbf{R}}$, and the segmented radial map estimate $\mathbf{\hat{R}}_{\hat{S}}$ is obtained through element-wise multiplication of $\hat{\mathbf{S}}$ and $\hat{\mathbf{R}}$. 

Our proposed DLT Surface Estimation method then takes $\mathbf{\hat{R}}_{\hat{S}}$ along with known 3D keypoints in the object frame to estimate the object surface points, where for each pixel ${p}_i$ in $\mathbf{\hat{R}}_{\hat{S}}$, the estimated object surface point is denoted as $\hat{\bar{p}}_i$, and the full set of estimated surface points is represented as $\widehat{\overline{\mathbf{X}}}$. 
Each 
$\hat{\bar{p}}_i$ in the object frame has a corresponding 
$p_i$ in the image frame,
both of which are indexed through the
associated image pixel.
To compute the full metric 6D pose of detected objects, we align $\mathbf{\widehat{\overline{X}}}$ with the input depth-derived 3D point cloud $\mathbf{I}_{\hat{S}}^D$ and estimate the 3D translation and rotation using a RANSAC-based Umeyama algorithm~\cite{umeyama, Fischler1981RandomSC}, subsequently refining this initial estimate with ICP~\cite{icp} to obtain our final 6D pose.

\subsection{Symmetric Keypoints}
\label{sec:Symmetric Keypoints}

Earlier keypoint-based methods ~\cite{Wu, He, Peng} assign a fixed keypoint order without considering object symmetries, potentially leading to inconsistent keypoint assignments when multiple valid symmetric configurations exist. This ambiguity is particularly problematic for objects with discrete symmetries, such as those 
with
$\pi$-radial symmetry, where 
the keypoints extracted in the image frame
can map to a variety of respective corresponding keypoints in the object frame,
depending on which equivalent symmetric
pose is selected.
Consequently, these methods struggle to consistently predict keypoints in a manner that preserves object symmetry, introducing regression ambiguities and errors which challenges the model to learn a stable representation.

Existing keypoint selection strategies, including bounding box keypoints~\cite{Wu}, farthest point sampling keypoints~\cite{He}, and KeyGNet keypoints~\cite{wu2024keygnet}, attempt to improve 
the accuracy and robustness of the overall
keypoint matching process.
Among these, KeyGNet mitigates symmetry-related ambiguities by learning optimal keypoint locations rather than relying on heuristic selection, 
thereby enhancing stability. However, even with these improved keypoint
locations,  
inconsistencies in ordering persist, as the keypoints still require correct indexing relative to the camera viewpoint for stable regression.

To address this challenge, we propose a symmetry-aware 
approach that enforces consistent keypoint
ordering during training
by dynamically adapting to object symmetries. The model estimates a four-channel radial map, 
where each channel encodes the radial distance of each image pixel to a predefined keypoint. Crucially, the structure of the radial map must remain consistent across symmetric object poses; otherwise, variations in keypoint ordering can lead to inconsistent regression targets and hinder generalization.

Unlike previous methods that rely on a fixed keypoint assignment, our approach dynamically reorders the radial channels based on the object's symmetry and its orientation relative to the camera. Specifically, we determine the ordering of keypoints by analyzing their proximity to the camera origin. 
The approach depends on designing
pairs of symmetric keypoints
which correspond to the symmetric object poses. 
These keypoints are generated by leveraging the Oriented Bounding Box (OBB), which provides a stable reference frame for structured keypoint placement. The OBB is computed using the Minimum Volume Enclosing Box algorithm~\cite{CHAN20011433}, ensuring a compact representation of the object. From the OBB, four side faces are identified 
manually,
each defined by four corner points, and the face centers are computed as their mean. 
The face normal vectors are determined using the cross-product of two independent edge vectors and subsequently normalized. The symmetric keypoints are then generated by translating these face centers along their normal vectors by a fixed offset distance 
\(d\). This ensures that all keypoints remain equidistant from the OBB center, preserving spatial symmetry.
A pseudocode representation of the above described algorithm 
is presented in Algorithm~\ref{alg:sym_keypoints} in Supp.
The method is semi-automatic, with Step 2 performed manually.
To our knowledge, this is the first time that
keypoint selection has been 
considered as an approach to address 
object symmetries.



\begin{figure}[t]
    \centering
    \includegraphics[height=4.5cm,width=0.85\linewidth]{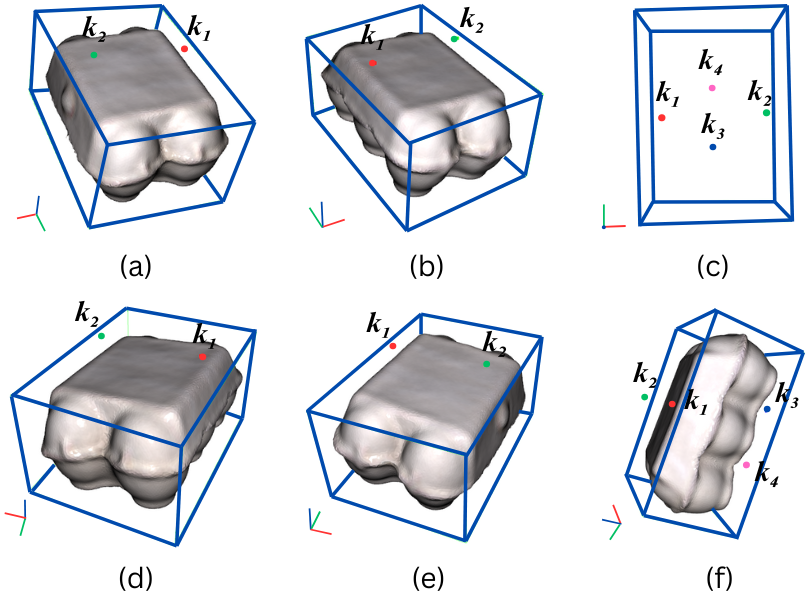}
    \caption{Eggbox keypoints under different rotations. (a) Original pose. (b) 180-degree Z-axis rotation, demonstrating symmetry. (c) Top-down view, highlighting symmetric keypoints $k_1$-$k_2$ (top) and $k_3$-$k_4$ (bottom). (d) -45-degree Z-axis rotation. (e) 135-degree Z-axis rotation, equivalent to a 180-degree rotation of (d). (f) Side view, showing the spatial distribution of keypoints $k_1$ to $k_4$.}
    \label{fig:sKpts}

\end{figure}

An example is shown in Fig.~\ref{fig:sKpts}, in which the object exhibits a $\pi$-radian symmetry between pairs of keypoints $k_1$–$k_2$ 
and $k_3$–$k_4$,
yielding a total of 4 possible orderings.
For the configuration in Fig.~\ref{fig:sKpts}(a), the four-channel radial map 
($\mathbf{R}_{S}$ of Fig.~\ref{fig:method}) is ordered as 
$[
\mathbf{R}_{S}^2,
\mathbf{R}_{S}^1,
\mathbf{R}_{S}^4,
\mathbf{R}_{S}^3]$ 
with the order determined by the keypoints' relative proximity to the camera, i.e. $k_2$ is closer 
to the camera origin than $k_1$, and $k_4$ is closer than $k_3$.
For Fig.~\ref{fig:sKpts}(b), the correct ordering is 
$[
\mathbf{R}_{S}^1,
\mathbf{R}_{S}^2,
\mathbf{R}_{S}^3,
\mathbf{R}_{S}^4]$ 
whereas in Fig.~\ref{fig:sKpts}(d), the order is 
$[
\mathbf{R}_{S}^1,
\mathbf{R}_{S}^2,
\mathbf{R}_{S}^4,
\mathbf{R}_{S}^3]$ 
while Fig.~\ref{fig:sKpts}(e) follows
order
$[
\mathbf{R}_{S}^2,
\mathbf{R}_{S}^1,
\mathbf{R}_{S}^3,
\mathbf{R}_{S}^4]$.
Ensuring a consistent radial map order relative to the camera viewpoint prevents inconsistencies in loss computation, stabilizing training and improving pose estimation under symmetric transformations,
as demonstrated in the ablation experiments of Sec.~\ref{sec:ablations}.

\section{Evaluation}
\subsection{Datasets and Evaluation Metrics}

Our method was trained and evaluated on BOP ~\cite{hodan2018bop} core datasets (LMO~\cite{Brachmann}, YCB~\cite{xiang2018posecnn},
TLESS~\cite{hodan2017tless}, TUDL~\cite{hodan2018bop}, ITODD~\cite{drost2017mvtec}, HB~\cite{kaskman2019homebreweddb}, and IC-BIN ~\cite{doumanoglou2016recovering}) along with LM~\cite{hinterstoisser2012model}, and we follow the standard BOP protocols in their use. For LM and LM-O,
we expand the training data using
BOP physics-based rendered data, 
which are fully synthetic training images, and PVNet-rendering~\cite{Peng} augmentation, which overlays objects from real images onto synthetic backgrounds. ITODD and HB have no real images in the training set, so
for training we instead used the real images in their validation sets, which were disjoint from their test sets. IC-BIN does not include any real training or validation images and is therefore trained on purely synthetic data.

Our primary evaluation metric is Average Recall ($AR$)~\cite{hodan2018bop}, 
which
evaluates pose estimation performance across three key components: Visible Surface Discrepancy ($AR_{VSD}$), Maximum Symmetry-Aware Surface Distance ($AR_{MSSD}$), and Maximum Symmetry-Aware Projection Distance ($AR_{MSPD}$). These components enable a fine-grained analysis of pose accuracy while effectively handling object symmetries. 
For completeness, we also report results using the widely adopted ADD(-S)~\cite{hinterstoisser2012model} metric for the LM and LM-O datasets, as well as the ADD-S and ADD(-S) AUC~\cite{xiang2018posecnn} metric for the YCB-V dataset.

\newcommand{\redding}{{\color{red}\ding{55}}}

\newcommand{\greencheck}{{\color{green}\checkmark}}

\begin{table*}[t]
\centering
\renewcommand{\arraystretch}{1}
\resizebox{\textwidth}{!}{
\begin{tabular}{lccccccccccccccc}
\toprule
\multirow{2}{*}{Method} & \multirow{2}{*}{Depth} & \multicolumn{1}{c}{1 Model} & \multicolumn{1}{c}{Refine-} & \multicolumn{2}{c}{ADD(-S)} & AUC(-S) & \multicolumn{9}{c}{Mean AR} \\
\cmidrule(lr){5-6} \cmidrule(lr){7-7} \cmidrule(lr){8-16}    
& & /Object & ment & LM & LM-O & YCB-V & LM & LM-O & YCB-V & T-LESS & TUDL & ITODD & HB & IC-BIN & Avg.\\
\midrule
PVNet~\cite{Peng} & \redding & \greencheck & \redding & 86.3 & 40.8 & 73.4 & -- & 0.575 & -- & -- & -- & -- & -- & -- & --\\
ZebraPose~\cite{su2022zebrapose} & \redding & \greencheck & \redding & -- & 76.9 & 85.3 & -- & 0.718 & 0.815 & 0.775 & 0.861 & 0.346 & 0.856 & -- & --\\
Pix2Pose~\cite{Park} & \redding & \greencheck & \redding & 72.4 & 32.0 & -- & -- & -- & -- & 0.295 & 0.420 & 0.134 & 0.446 & 0.226\\
CosyPose~\cite{labbe2020} & \redding & \redding  & \greencheck & -- & -- & 84.5 &  -- & 0.633 & 0.821 & 0.728 & 0.823 & 0.216 & 0.656 & 0.583 & 0.637\\
PoseCNN~\cite{xiang2018posecnn} & \greencheck & \redding & \greencheck  &  -- & 78.0 & 79.3 & -- & -- & -- & -- & -- & -- & -- & -- & --\\ 
PVN3D~\cite{He} & \greencheck & \greencheck & \greencheck & 99.4 & 63.3 & 92.3 & -- & -- & -- & -- & -- & -- & -- & -- & --\\
RCVPose~\cite{Wu} & \greencheck & \greencheck & \greencheck & 99.7 & 71.1 & \underline{97.2} & 0.799 & 0.749 & 0.859 & -- & -- & -- & -- & -- & --\\
RCVPose3D~\cite{wu2022keypoint} & \greencheck & \redding & \greencheck  & -- & 74.5 & 96.6 & -- & 0.729 & 0.843 & 0.708 & 0.966 & 0.536 & 0.863 & \underline{0.733} & 0.768\\
FFB6D~\cite{he2021ffb6d} & \greencheck & \greencheck & \redding & 99.7 & 66.2 & 92.7 & 0.773 & 0.687 & 0.758 & -- & -- & -- & -- & -- & --\\
IRPE~\cite{Jin2024IRPEIR} & \greencheck & \redding & \greencheck  & 98.8 & 85.4 & 90.6 & -- & -- & -- & -- & -- & -- & -- & -- & --\\
SurfEmb~\cite{Haugaard} & \greencheck & \redding & \greencheck & -- & -- & -- & -- & 0.758 & 0.824 & 0.833 & 0.933 & 0.498 & 0.867 & 0.656 & 0.767\\
HiPose~\cite{lin2024hipose} & \greencheck & \greencheck & \redding &  -- & 89.6 & 95.5 & -- & \underline{0.799} & 0.907 & 0.833 & -- & -- & -- & -- & --\\
EPRO-GDR~\cite{10896093}  & \greencheck & \redding & \greencheck  & -- & 88.7 & 92.6 & -- & 0.786 & 0.844 & 0.765 & -- & 0.412 & -- & -- & --\\
CIR~\cite{lipson2022coupled} & \greencheck & \redding & \greencheck & -- & -- & -- & -- & 0.734 & 0.893 & 0.715 & \underline{0.968} & 0.381 & 0.757 & 0.676 & 0.741\\
GenFlow~\cite{Moon2024GenFlow} & \greencheck & \redding & \greencheck  & -- & -- & -- & -- & 0.744 & 0.884 & 0.780 & 0.924 & 0.647 & 0.916 & 0.651 & 0.792\\
DFTr~\cite{zhou2023deep} & \greencheck & \greencheck & \greencheck & \underline{99.8} & 77.7 & 94.8 & -- & -- & -- & -- & -- & -- & -- & -- & --\\
GDRNPP~\cite{liu2025gdrnpp} & \greencheck & \greencheck & \greencheck & -- & \textbf{93.7} & -- & -- & 0.775 & \textbf{0.921} & \textbf{0.874} & 0.966 & \underline{0.679} & \underline{0.926} & 0.722 & \underline{0.837}\\
GDRNPP~\cite{liu2025gdrnpp}
& \greencheck & \greencheck & \redding & --& -- & -- & --
& 0.713 & 0.825 & 0.786 & 0.831 & 0.448 & 0.869 & 0.623 & 0.728 \\

DLTPose (Ours, w/o ICP)
& \greencheck & \greencheck & \redding & 99.0 & 83.6 & 94.3 & 0.825
& 0.725 & 0.781 & 0.778 & 0.968 & 0.626 & 0.860 & 0.698 & 0.777\\

DLTPose (Ours) & \greencheck & \greencheck & \greencheck & \textbf{99.9} & \underline{91.0} & \textbf{97.4} & \textbf{0.865} & \textbf{0.800} & \underline{0.909} & \underline{0.839} & \textbf{0.972} & \textbf{0.722} & \textbf{0.929} & \textbf{0.839} & \textbf{0.839}\\
\bottomrule
\end{tabular}}
\caption{Comparison of \textsc{DLTPose} with recent state-of-the-art methods on the BOP core datasets and LM dataset. Avg. is only taken over the 7 Core BOP datasets (excluding LM). Along with accuracy metrics, we report the use of depth input and per-object model training.
}
\label{tab:bop_ar}
\end{table*}

\subsection{Implementation}
Prior to training, 
RGB images are normalized to the range [0,1]. Segmented depth maps are independently normalized using their minima and maxima to ensure consistent depth scaling across different scenes. Instead of computing radial distances directly from the depth field, radial maps are derived from the transformed object mesh and keypoints, ensuring consistency in distance estimation. 
For asymmetric objects, object frame keypoints are chosen automatically using KeyGNet,
whereas for symmetric objects, keypoints are chosen semi-automatically using the algorithm described in Sec.~\ref{sec:Symmetric Keypoints}.
These radial maps are expressed in decimeter units, 
allowing the network to operate within a stable numerical range which improves  accuracy~\cite{Wu}.

For optimization, we use the Adam optimizer~\cite{adam} with an initial learning rate of 
1e-3,
following the loss functions in Eqs.~\eqref{eq:l_r}--\eqref{eq:l_sc}. The learning rate is dynamically adjusted using a Reduce-on-Plateau strategy, where it is reduced by a factor of 0.1 upon stagnation. The method is implemented in PyTorch~\cite{pytorch}, with each object being trained on a separate model
for 200-250 epochs, using a batch size of 32.

Additionally, we train a custom Mask R-CNN~\cite{maskrcnn} model,
pretrained on ImageNet~\cite{imgnet}, to detect and segment the target objects in the scene. During training, RGB images are shifted and scaled to match the ImageNet mean and standard deviation. Models are trained for $\sim\!\!50$ epochs with batch size 4 and initial learning rate
1e-4. 
For convenience, we train one model per object,
although we found performance was
similar when training one model per dataset.
In particular, for LM-O,
Mean ADD(-S) was the same for both training modes
(Table~\ref{tab:SISO_MIMO_performance})
while Mean AR was comparable
(Table~\ref{tab:SISO_MIMO_performance}).
All training is conducted on a server equipped with an Intel Xeon 5218 CPU and three RTX8000 GPUs or three A100 GPUs.
The ResNet-152 backbone used in our framework contains approximately 165.3M parameters.

\subsection{{Results}}

Table~\ref{tab:bop_ar} presents a comprehensive comparison of DLTPose with recent state-of-the-art methods on the BOP core datasets and LM dataset across multiple evaluation metrics. In addition to performance metrics, the table also reports whether or not each method uses depth input
and trains separate models per object. DLTPose achieves the highest mean performance across all the datasets, while also leading in ADD(-S) and AUC(-S) metrics. Detailed per-object results on ADD(-S) and AUC(-S) metrics are provided in the Supplementary Material (Tables~\ref{tab:performance_lm}--\ref{tab:performance_tless}).

\noindent \textbf{Runtime Analysis. }
Inference experiments were conducted on a system with an Intel Core i9 13th Gen CPU and a single NVIDIA P100 GPU.
Our average object pose estimation time across the datasets is 0.627 seconds per image, including ICP refinement.
The average 2D detection time with Mask R-CNN~\cite{maskrcnn} 
is 0.067 seconds.
A comparison with the runtimes of other methods is shown in Figure~\ref{fig:runtime} of the Supplementary Material. This shows our runtime as the median among all methods.

\begin{figure}[t]
    \centering
\includegraphics[height=4.8cm, width=0.9\linewidth]{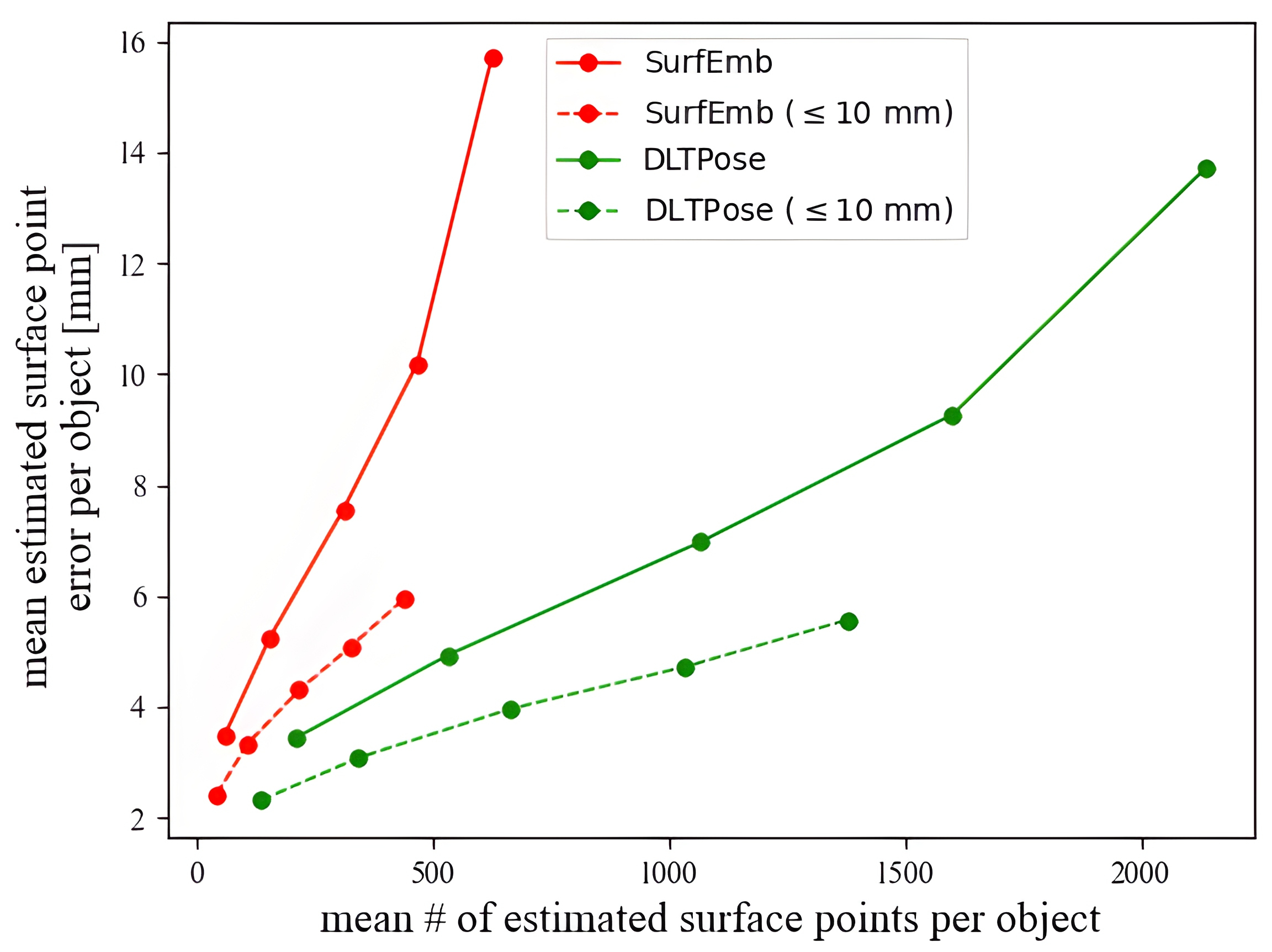}
    \caption{Mean error vs. mean number of estimated surface points per object, for SurfEmb (red) and DLTPose (green). Solid lines average over all estimated points; dashed lines only include points with errors $\le 10$ mm. 
    }
    \label{fig:error_surface}
\end{figure}

\noindent \textbf{Surface Estimation Accuracy Experiments.}
We conducted two experiments that show the improved
accuracies and densities of surface points estimated
with our DLT method,
and its benefit on pose estimation.
The first 
compares the accuracies
and number of surface points generated by leading dense approach SurfEmb~\cite{Haugaard} with our DLT approach, comparing them against ground truth values. 
Ground truth 2D-3D correspondences are obtained from the 3D mesh model, foreground mask, and object pose for each object in the LM-O dataset. From the known object pose, the 3D mesh is transformed into the camera frame. Each foreground point in the transformed mesh is then projected onto the image plane using the camera intrinsics, producing a ground truth set
in which each 2D image pixel indexes a corresponding 3D object surface point.

To assess accuracy of the estimates, we generated corresponding sets for both SurfEmb and our DLT approach and computed the point estimation error by comparing them against the ground truth correspondences. 
The errors between ground truth and estimated points for each method was measured across different percentile levels (i.e. top $10^{th}, 25^{th}, 50^{th}, 75^{th},$ and $100^{th}$ percentile) of the estimated correspondences,
sorted by increasing error. 
The results 
are shown 
in Fig.~\ref{fig:error_surface},
each curve plotting values from the smallest 
($10^{th}$, leftmost)
to the largest
($100^{th}$, rightmost)
percentile.
The solid curves include all estimated points,
whereas the dashed curves include only points
that
lie within 10 mm of their ground truth corresponding values.
These results demonstrate that the DLT approach outperforms SurfEmb in both accuracy and density of  surface point estimates. As the number of estimated surface points increases, SurfEmb’s error rises sharpely, indicating higher deviations from the ground truth, whereas the DLT approach maintains lower errors across all cases. When filtering correspondences, and considering errors $\leq 10$ mm, both methods show improved accuracy, 
with the DLT approach retaining a greater number of higher accuracy surface points. Overall, the DLT  approach achieves 
both increased accuracy and increased point count.


\noindent \textbf{Impact of Surface Noise on Pose Estimation Accuracy.}
A second experiment showed the impact of higher accuracy surface estimates on pose estimation. 
The estimated object surface points were perturbed by zero-mean Gaussian noise with varying standard deviations,
and the $AR$ metric was evaluated
for all LM-O objects following pose estimation,
with the frontend RANSAC-enabled Umeyama used during inference. 
The results plotted in Figure~\ref{fig:error_surface_pose}
show that
as noise increases, performance degrades across metrics MSSD, MSPD, and $AR$, indicating that accurate surface point estimation impacts pose estimation accuracy.

\begin{figure}[t]
    \centering
\includegraphics[height=4.8cm, width=0.9\linewidth]{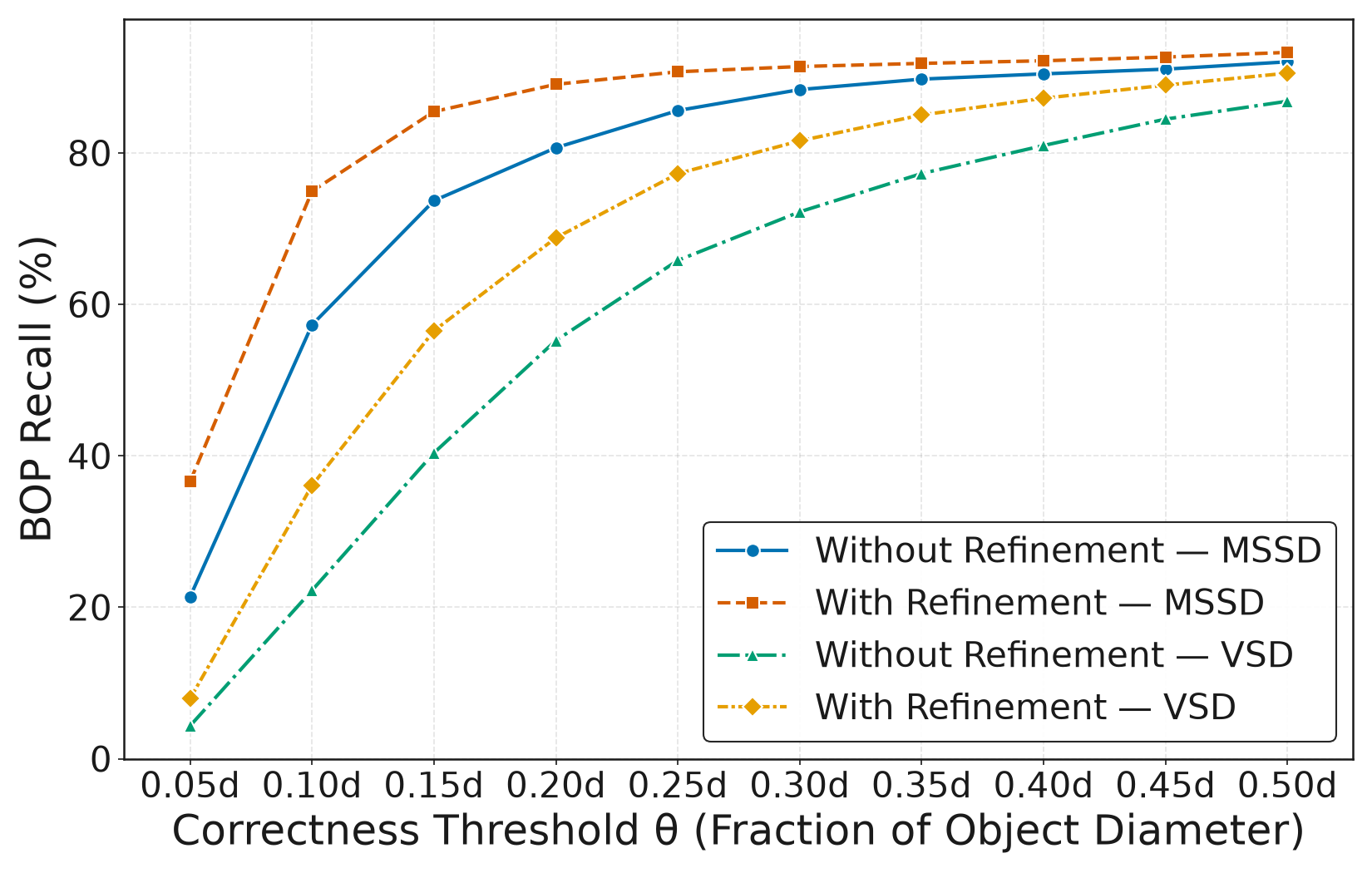}
    \caption{MSSD and VSD recall versus correctness threshold $\theta$ on LM-O, with and without ICP refinement.
    }
    \label{fig:threshold_sweep}
\end{figure}

\noindent \textbf{Impact of AR Sensitivity on Refinement.}
BOP Average Recall averages performance across correctness thresholds $\theta$, masking where refinement provides the greatest benefit. As shown in Fig.~\ref{fig:threshold_sweep}, sweeping $\theta$ from $0.05d$ to $0.50d$ on LM-O shows that ICP primarily improves precision rather than recovering failed poses. At the loosest $\theta = 0.50d$ threshold, MSSD recall increases only from $92.0\%$ to $93.3\%$ ($+1.3$ points), indicating that most initial poses are already approximately correct. In contrast, at $\theta = 0.10d$, ICP improves MSSD from $57.3\%$ to $75.0\%$ ($+17.7$ points) and VSD from $22.2\%$ to $36.1\%$ ($+13.9$ points), while gains beyond $0.30d$ fall below $3$ points. Overall, ICP improves $\mathrm{AR}_{\mathrm{MSSD}}$ from $0.770$ to $0.838$ and $\mathrm{AR}_{\mathrm{VSD}}$ from $0.590$ to $0.680$, confirming that it mainly converts coarse-but-correct estimates into more accurate poses.

\noindent \textbf{Ablations.}
\label{sec:ablations}
We conducted additional experiments to evaluate: (i) the impact of symmetric keypoints, (ii) the effect of the pseudo-symmetric loss during training, (iii) the performance difference between single-object vs. all-object models, (iv) the influence of using object segmentations versus bounding boxes during training, and (v) the effect of varying the number of keypoints (4, 8, 12, 16). The details are described in the Supplementary Material, Sec.~\ref{sec:ablation_experiments}. 
In summary, (i) symmetric keypoints significantly enhance ADD-S and AR metrics by explicitly addressing pose ambiguities for symmetric objects, (ii) incorporating pseudo-symmetric loss improves ADD(-S) scores by refining radial map regression accuracy, (iii) single-object models achieve higher AR compared to a unified all-object model, as they effectively learn object-specific geometric constraints, (iv) Fewer keypoints (especially 4) achieves superior or comparable accuracy by reducing per-channel radial regression errors, and (v) segmented inputs consistently outperform bounding boxes due to more accurate localization of object boundaries, yielding precise radial predictions. 
\section{Conclusion}
We have presented DLTPose, which estimates 3D surface points from a minimal set of four keypoints using a novel DLT formulation. The surface points are shown to be highly accurate, which leads to improved pose estimation accuracy. In addition, we present a symmetric keypoint ordering method that dynamically orders keypoints, thereby reducing ambiguities of regressed values during training.
Our method achieves state-of-the-art performance on the benchmark BOP core datasets, outperforming recent methods, including leading dense methods such as SurfEmb. Its performance is particularly strong for symmetric objects, where symmetric keypoint ordering  improves accuracy.

A potential limitation of DLTPose is that the DLT formulation requires non-coplanar keypoints, although
in practice keypoints are easily validated as being non-coplanar during preprocessing. A further limitation is that the approach employs distinct keypoint strategies for asymmetric (automatically with KeyGNet) and symmetric (semi-automatically as described in Sec.~\ref{sec:Symmetric Keypoints}) objects. 
%
Future work will focus on methods to 
further improve the accuracy of surface estimation, which will in turn  improve pose estimation accuracy. 
We will also explore other approaches for automatically
generating 
and
dynamically ordering 
symmetric keypoints  
during training.

{   \small
    \bibliographystyle{ieeenat_fullname}
    \bibliography{main}
}

\clearpage
\clearpage
\setcounter{page}{1}
\maketitlesupplementary
\setcounter{section}{0}
\renewcommand{\thesection}{S.\arabic{section}}
\renewcommand{\thesubsection}{S.\arabic{section}.\arabic{subsection}}

\setcounter{table}{0}
\renewcommand{\thetable}{S.\arabic{table}}
\setcounter{figure}{0}
\renewcommand{\thefigure}{S.\arabic{figure}}
\setcounter{equation}{0}
\renewcommand{\theequation}{S.\arabic{equation}}

\section{Overview}
This document provides additional details and experiments supporting our main work. Section~\ref{sec:DLT Derivation} presents the full derivation of the Direct Linear Transform (DLT) formulation. Section~\ref{sec:supp_skpts} details our symmetric keypoint generation approach using Oriented Bounding Boxes (OBB). Section~\ref{sec:common_points} evaluates surface estimation accuracy by comparing error distributions for common pixel-aligned correspondences between our method and SurfEmb. Section~\ref{sec:ablation_experiments} covers ablation studies on symmetric keypoints, pseudo-symmetric loss, and surface estimation noise. Section~\ref{sec:ablation_results} provides per-object accuracy results for LINEMOD, Occlusion LINEMOD, and YCB-Video, including ADD(-S) and AUC scores, along with qualitative pose recovery examples.

\section{DLT Derivation}
\label{sec:DLT Derivation}
Let 
$\bar{p}=(\bar{x},\bar{y},\bar{z})$ 
be an object surface point
and let
$\bar{k}=(\bar{x}_k,\bar{y}_k,\bar{z}_k)$
be a keypoint,
with
$\bar{p}$
and
$\bar{k}$
both described in a common reference frame, such as the object frame
without loss of generality.
Expanding Eq.~\ref{eq:r_ij}, and for simplicity dropping the subscripts on 
$\bar{k}_j$ and $r_j$,
the square of the radial distance $r$ between 
$\bar{p}$
and
$\bar{k}$
is expressed as:
\begin{equation}
\label{eq:r2}
(\bar{x}-\bar{x}_k)^{2}
+
(\bar{y}-\bar{y}_k)^{2}
+
(\bar{z}-\bar{z}_k)^{2}
=
r^2.
\end{equation}
Expanding the terms of Eq.~(\ref{eq:r2}) gives:
\begin{equation}
\label{eq:r2exp}
\bar{x}^2
-2 \bar{x} \bar{x}_k
+ \bar{x}_k^2
+
\bar{y}^2
-2 \bar{y} \bar{y}_k
+ \bar{y}_k^2
+
\bar{z}^2
-2 \bar{z} \bar{z}_k
+ \bar{z}_k^2
-
r^2
= 0,
\end{equation}
and further rearranging terms yields:
\begin{equation}
-2 \bar{x} \bar{x}_k
-2 \bar{y} \bar{y}_k
-2 \bar{z} \bar{z}_k
+
(
\bar{x}^2
+
\bar{y}^2
+
\bar{z}^2
)
+
(
\bar{x}_k^2
+
\bar{y}_k^2
+
\bar{z}_k^2
-
r^2
)
= 0.
\end{equation}

All known (i.e. constant or measured) quantities
can be collecting into a left vector $A$, and multiplied with a right vector $X$ containing all unknown quantities, as follows:
\begin{equation}
\begin{aligned}
\begin{bmatrix}
-2\bar{x}_k &
-2\bar{y}_k &
-2\bar{z}_k &
1 &
\left(\bar{x}_k^2 +
\bar{y}_k^2 +
\bar{z}_k^2 -
r^2\right)
\end{bmatrix}
\\[0.5em]
\cdot
\begin{bmatrix}
\bar{x} \\
\bar{y} \\
\bar{z} \\
\left(\bar{x}^2 +
\bar{y}^2 +
\bar{z}^2\right) \\
1
\end{bmatrix}
= 0
\end{aligned}
\end{equation}

Finally, we simplify the notation to
$|\!|\bar{k}|\!|^2 = \bar{x}_k^2 
+
\bar{y}_k^2 
+
\bar{z}_k^2$
and 
$|\!|\bar{p}|\!|^2 = \bar{x}^2 
+
\bar{y}^2 
+
\bar{z}^2$
and stack a series of $N_k$
such rows into matrix $A$ to yield Eq.~\ref{eq:matrix3}:
\begin{equation*}
\begin{bmatrix}
\!\! -2\bar{x}_{k_1} \rs 
& \rss -2\bar{y}_{k_1} \rs
& \rs -2\bar{z}_{k_1} \rs
& \rss 1 
& \rss
(|\!|\bar{k}_1|\!|^2 \!-\! \hat{r}_{1}^2) \\
\vdots & \vdots & \vdots & \rss \vdots & \vdots \\
\! -2\bar{x}_{k_{N_k}} \rs
& -2\bar{y}_{k_{N_k}} \rs
& -2\bar{z}_{k_{N_k}} \rss
& \rss 1 
& \rss
(|\!|\bar{k}_{N_k}|\!|^2 \!-\! \hat{r}_{N_k}^2) \\
\end{bmatrix}
\begin{bmatrix}
\bar{x} \\
\bar{y} \\
\bar{z} \\
|\!|\bar{p}|\!|^2 \\
1
\end{bmatrix}
= 0.
\end{equation*}

\begin{figure}[ht]
    \centering
\includegraphics[width=1.0\linewidth]{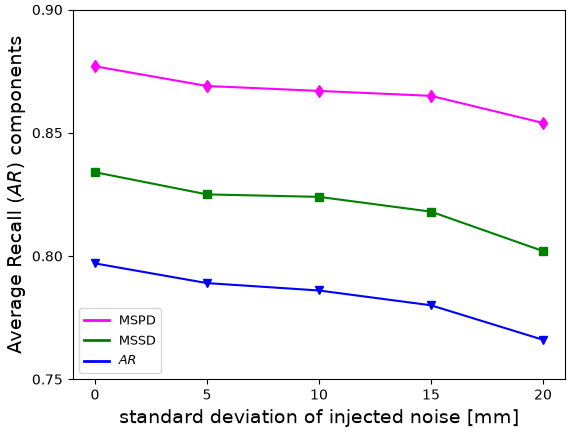}
    \caption{Average recall components (MSSD, MSPD, and $AR$) vs. standard deviation of injected noise on the estimated LM-O object surface points.
    }
    \label{fig:error_surface_pose}
\end{figure}

\begin{figure}[ht]
    \centering
\includegraphics[width=1.0\linewidth]{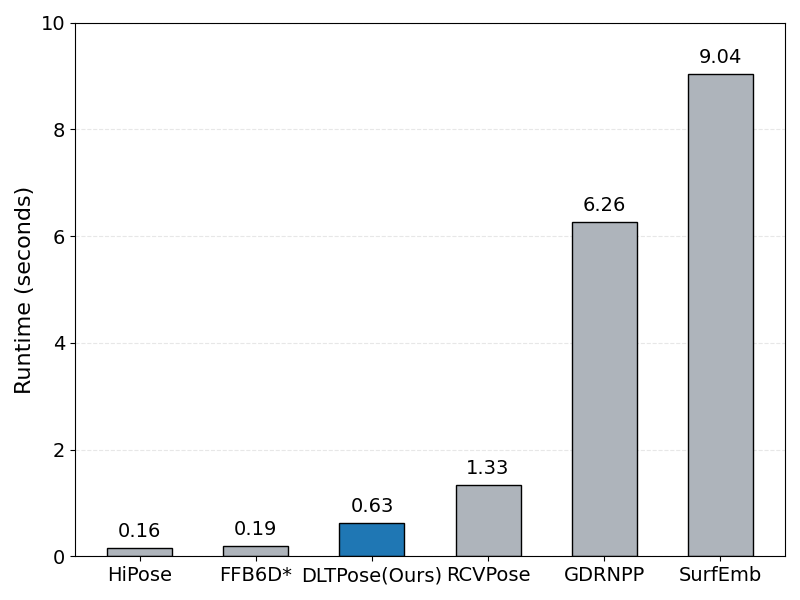}
    \caption{Runtime comparison of our method with prior ap-
proaches averaged over LM-O, YCB-V and T-LESS datasets.
$^{\ast}$FFB6D is averaged over LM-O and YCB-V only as T-LESS results were not
provided for this method.
    }
    \label{fig:runtime}
\end{figure}

\section{Symmetric Keypoints Generation}
\label{sec:supp_skpts}

Thrun and Wegbreit observe that the symmetry present in objects can be described by two orthogonal reflection planes~\cite{thrun2005shape}. We evaluate symmetry directly on the object’s Oriented Bounding Box computed via the Minimum Volume Enclosing Box algorithm~\cite{CHAN20011433}. The OBB induces three mutually orthogonal mid-planes. For each plane, one half of the object is reflected across the plane and a bidirectional Chamfer distance is computed between the reflected and original halves to quantify reflective similarity. The planes are ranked by this score and the two highest scoring planes are retained as candidate symmetry planes. The four OBB side faces associated with these two planes are then selected in a principled manner for symmetric keypoint generation. This procedure replaces ad hoc face selection with a deterministic, data-driven criterion aligned with the object’s dominant symmetries.

By identifying four primary side faces and computing their centers, the approach ensures a structured and uniform distribution of keypoints. The normal vectors of these faces, obtained through the cross-product of independent edge vectors, guide the placement process. Each center is shifted along its normal by a fixed offset \d, preserving equal spacing relative to the OBB center. This technique enforces stable keypoint assignments across symmetric object poses and the accompanying pseudocode in Algorithm~\ref{alg:sym_keypoints} provides a clear implementation framework for integrating this approach into practical applications.

\begin{algorithm}[h]
\caption{Symmetric Keypoint Generation from Oriented Bounding Box (OBB)}
\label{alg:sym_keypoints}
\begin{algorithmic}[1]
\Require Object mesh $\mathcal{M}$, Offset distance $d$
\Ensure Symmetric keypoints $\mathcal{K} = \{k_1, k_2, k_3, k_4\}$

\State \textbf{Step 1: Extract Oriented Bounding Box (OBB)}
\State Compute OBB from $\mathcal{M}$ using Minimum Volume Enclosing Box.
\State Retrieve 8 corner points $\mathcal{Q} = \{q_0, q_1, \dots, q_7\}$.

\State \textbf{Step 2: Define Side Faces}
\State $F_1 = \{ q_0, q_1, q_5, q_4 \} 
= \{q_0^1,q_1^1,q_2^1,q_3^1\}$
\State $F_2 = \{ q_2, q_3, q_7, q_6 \} 
= \{q_0^2,q_1^2,q_2^2,q_3^2\}$
\State $F_3 = \{ q_0, q_2, q_6, q_4 \} 
= \{q_0^3,q_1^3,q_2^3,q_3^3\}$
\State $F_4 = \{ q_1, q_3, q_7, q_5 \} 
= \{q_0^4,q_1^4,q_2^4,q_3^4\}$

\State $\cal{K} = \{\}$
\ForAll{$F_i \in \{F_1, F_2, F_3, F_4\}$}
    \State \textbf{Step 3: Compute Face Center}
    \State Compute face center: $f_i = \frac{1}{4} \sum{\{q_j^i\}_{j=0}^{3}}$ 
    \State \textbf{Step 4: Compute Face Normal Vector}
    \State Compute edge vectors: $\vec{e}_1\!=\!q_1^i - q_0^i, \;\vec{e}_2\!=\!q_3^i - q_0^i$
     \State Compute normal vector: $\vec{n}_i = \frac{\vec{e}_1 \times \vec{e}_2}
     {||\vec{e}_1 \times \vec{e}_2||}$

    \State \textbf{Step 5: Compute Symmetric Keypoints}
    \State Compute symmetric keypoint: $k_i = f_i + d \cdot \vec{n}_i$
    \State $\mathcal{K} = \mathcal{K} + k_i$
\EndFor

\end{algorithmic}
\end{algorithm}

\section{Surface Estimation, Common Points}
\label{sec:common_points}
This experiment evaluates the accuracy of surface point estimates generated by SurfEmb~\cite{Haugaard} and our DLT approach, focusing specifically on those common pixels in the scene where both methods provided corresponding 3D surface estimates. The objective is to compare the estimation error for overlapping correspondences, ensuring a fair assessment of each method’s accuracy in shared regions.

\begin{figure}[h]
    \centering
  \includegraphics[height=5.4cm,width=1.0\linewidth]{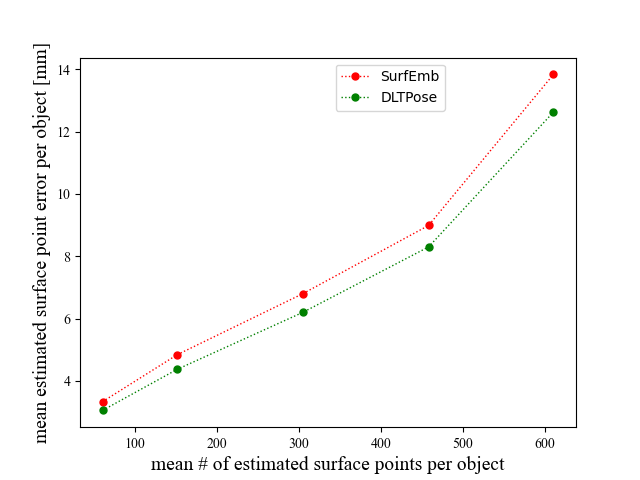}
    \caption{Mean error vs. mean number of estimated surface points per object, for SurfEmb (red) and DLTPose (green). A common set of points are evaluated for each method at each percentile level ($10^{th}, 25^{th}, 50^{th}, 75^{th}, 100^{th}$), sorted by increasing error. 
    }
    \label{fig:surf_estimates_common}

\end{figure}

The mean error per object in LMO is computed for these common pixels, and the results are visualized in Fig.~\ref{fig:surf_estimates_common}. The comparison shows that the DLT approach consistently achieves lower estimation error than SurfEmb across all density levels, demonstrating superior surface point accuracy even in overlapping regions. These findings reinforce that the DLT approach offers improved geometric consistency and more reliable surface reconstruction, ultimately leading to better pose estimation.

\section{Ablation Experiments}
\label{sec:ablation_experiments}


\noindent\textbf{Effect of Symmetric Keypoints on Pose Estimation.}
This experiment evaluates the impact of the proposed symmetric keypoint framework described in Sec.~\ref{sec:Symmetric Keypoints} on 6DoF pose estimation for objects with rotational symmetries. We compare 
the 
pose estimation results on the symmetric eggbox (from LM-O)
and bowl (from YCB-V) objects,
using the ADD-S and AR metrics.
For each object, the keypoints were selected using 
either
KeyGNet~\cite{wu2024keygnet} 
or
our symmetric keypoint framework.
%

\begin{table}[t]
\centering
\renewcommand{\arraystretch}{1.2}
\resizebox{\linewidth}{!}
{
\begin{tabular}{lcccc}
\toprule
\multirow{3}{*}{Object} & \multicolumn{2}{c}{KeyGNet~\cite{wu2024keygnet} keypoints} & \multicolumn{2}{c}{Symmetric keypoints} \\
\cmidrule(lr){2-3} \cmidrule(lr){4-5} 
& ADD-S & $AR$ & ADD-S & $AR$\\ 
\midrule
LM-O eggbox & 97.8 & 0.544 & \textbf{98.6} & \textbf{0.691} \\
YCB-V bowl & 97.9 & 0.715 & \textbf{100.0} & \textbf{0.893}  \\
YCB-V wood\_block & \textbf{100.0} & 0.964 &  \textbf{100.0} & \textbf{0.988}  \\
\bottomrule
\end{tabular}}
\caption{Comparison of ADD-S and $AR$ metrics for three symmetric objects, using keypoints selected using either KeyGNet~\cite{wu2024keygnet} or our symmetric keypoint generation method 
(Sec.~\ref{sec:Symmetric Keypoints}).}
\label{tab:keygnet_obb_performance}
\end{table}

Table~\ref{tab:keygnet_obb_performance} shows that our symmetric keypoints consistently improve performance. On eggbox, ADD-S increases from 97.8\% to 98.6\%, while AR improves from 0.544 to 0.691. Similarly, bowl sees an increase in ADD-S from 97.9\% to 100.0\%, with AR rising from 0.715 to 0.893. The YCB-V wood block shows further improvement, as AR improves from 0.964 to 0.988 when using our symmetric keypoints with ADD-S maintaining 100.0\%. 
These results demonstrate that explicitly enforcing the proposed symmetric keypoint framework reduces regression ambiguities and improves pose estimation accuracy for symmetric objects.
This is especially evident from the improvement in the $AR$ scores,
which is more discriminating
than the coarser ADD(-S) metric.

\noindent\textbf{Effect of Pseudo-Symmetric Loss on Training.}
This ablation study examines the effect of different loss configurations on model performance by comparing training results using $\mathcal{L}_{R}$ (Eq.~\ref{eq:l_r}), a weighted combination of $\mathcal{L}_{R}$ and $\mathcal{L}_{C}$ ($\lambda_1\cdot\mathcal{L}_{R}+\lambda_2\cdot\mathcal{L}_{C}$, Eq.~\ref{eq:l_c}, with $\lambda_1=0.7, \lambda_2=0.3$), and the full loss $\mathcal{L}_{\text{total}}$ (Eq.~\ref{eq:t_loss}). Table~\ref{tab:diff_loss_performance} presents ADD(-S) results on LM-O across multiple objects.

\begin{table}[h]
\centering
\tiny
\resizebox{\linewidth}{!}{
\begin{tabular}{l|lccc}
\toprule
\multirow{3}{*}{\rotatebox{90}{Loss}} & $\mathcal{L}_{R}$ (Eq.~\ref{eq:l_r}) & \checkmark & \checkmark & \checkmark \\
 & $\mathcal{L}_{C}$ (Eq.~\ref{eq:l_c}) & -- & \checkmark & \checkmark \\
 & $\mathcal{L}_{P}$ (Eq.~\ref{eq:l_sc}) & -- & -- & \checkmark \\
\midrule
\multirow{8}{*}{\rotatebox{90}{Objects}}
& ape & 85.5 & \underline{86.5} & \textbf{87.1} \\
& can & 94.2 & \underline{96.2} & \textbf{96.8} \\
& cat & 67.5 & \underline{69.5} & \textbf{71.0} \\
& driller & \underline{95.5} & 95.1 & \textbf{97.3} \\
& duck & 83.4 & \underline{84.5} & \textbf{88.5} \\
& eggbox$\ast$ & 96.9 & \underline{97.6} & \textbf{97.8} \\
& glue$\ast$ & \underline{90.8} & 90.7 & \textbf{91.1} \\
& holepuncher & 93.6 & \underline{93.9} & \textbf{97.5} \\
\midrule
& Mean & 88.4 & \underline{89.3} & \textbf{91.0} \\ 
\bottomrule
\end{tabular}
}
\caption{Comparison of ADD(-S) metric on LM-O for models trained with KeyGNet keypoints and three loss function configurations.
}
\label{tab:diff_loss_performance}
\end{table}
The results indicate that integrating pseudo-symmetric loss ($\mathcal{L}_{\text{total}}$) improves mean ADD(-S) by +2.6\%. This demonstrates that enforcing pseudo-symmetry during training enhances radial map regression, leading to more accurate surface estimates and improved pose estimation accuracy.

\hfill \break
\noindent\textbf{Effect of Training a Single Model on the Entire LM-O Dataset.}
In this ablation study, we investigate the consequences of training one unified model on the LM-O dataset, which contains both symmetric and asymmetric objects. We optimize the model using the total loss $\mathcal{L}_{\text{total}}$ (Eq.~\ref{eq:t_loss}) and assess its performance via the ADD-S and AR metrics. Table~\ref{tab:SISO_MIMO_performance} reports the ADD(-S) scores on LM-O for multiple objects. Training a single unified model on the full dataset yields same ADD(-S) performance to training separate models for each individual object.

\begin{table}[h]
\centering
\resizebox{1.0\linewidth}{!}{
\begin{tabular}{lcccc}
\toprule
\multirow{1}{*}{Object} & \multicolumn{1}{c}{One Model per Object} & \multicolumn{1}{c}{One Model per Dataset} \\ 
\midrule
ape & \textbf{87.1} & \underline{86.5} \\
can & \underline{96.8} & \textbf{98.2} \\
cat & \underline{71.0} & \textbf{71.8} \\
driller & \textbf{97.3} & \underline{96.1} \\
duck & \textbf{88.5} & \underline{88.0} \\
eggbox$\ast$ & \textbf{98.6} & \underline{97.6} \\
glue$\ast$ & \textbf{91.1} & \underline{90.7} \\
holepuncher & \underline{97.5} & \textbf{98.6} \\
\midrule
Mean & \textbf{91.0} & \textbf{91.0} \\ 
\bottomrule
\end{tabular}}
\caption{Comparison of ADD(-S) metric, on One Model per Object and One Model per Dataset, on LM-O for models trained with KeyGNet and symmetric keypoints using the total loss $\mathcal{L}_{\text{total}}$ (Eq.~\ref{eq:t_loss}).
}
\label{tab:SISO_MIMO_performance}
\end{table}

\begin{table}[h]
\centering
\renewcommand{\arraystretch}{1.2}
\Large
\resizebox{\linewidth}{!}{
\begin{tabular}{lcccccccc}
\toprule
\multirow{2}{*}{Object} & \multicolumn{4}{c}{One Model per Object} & \multicolumn{4}{c}{One Model per Dataset} \\ 
\cmidrule(lr){2-5} \cmidrule(lr){6-9}  
& $\textit{AR}_{VSD}$ & $\textit{AR}_{MSSD}$ & $\textit{AR}_{MSPD}$ & $\textit{Mean AR}$ & $\textit{AR}_{VSD}$ & $\textit{AR}_{MSSD}$ & $\textit{AR}_{MSPD}$ & $\textit{Mean AR}$ \\ 
\midrule
ape & \underline{0.638}  & \textbf{0.830} & \textbf{0.912} & \textbf{0.795} & \textbf{0.645} & \underline{0.822} & \underline{0.906} & \underline{0.791} \\
can & \underline{0.737} & \underline{0.914} & \underline{0.921} & \underline{0.857} & \textbf{0.744} & \textbf{0.921} & \textbf{0.926} & \textbf{0.864} \\
cat & \textbf{0.568} & \textbf{0.713} & \textbf{0.797} & \textbf{0.693} & \underline{0.552} & \underline{0.697} & \underline{0.788} & \underline{0.679} \\
driller & \textbf{0.788} & \textbf{0.937} & \textbf{0.919} & \textbf{0.881} & \underline{0.773} & \underline{0.922} & \underline{0.901} & \underline{0.865} \\
duck & \underline{0.725} & \underline{0.808} & \textbf{0.903} & \textbf{0.812} & \textbf{0.728} & \textbf{0.812} & \underline{0.896} & \textbf{0.812} \\
eggbox$\ast$ & \textbf{0.560} & \textbf{0.722} & \textbf{0.790} & \textbf{0.691} & \underline{0.467} & \underline{0.593} & \underline{0.668} & \underline{0.576} \\
glue$\ast$ & \underline{0.648} & \underline{0.842} & \underline{0.851} & \underline{0.780} & \textbf{0.677} & \textbf{0.848} & \textbf{0.864} & \textbf{0.796} \\
holepuncher & \underline{0.738} & \underline{0.901} & \textbf{0.932} & \textbf{0.859} & \textbf{0.741} & \textbf{0.908} & \underline{0.930} & \textbf{0.859} \\
\bottomrule
\end{tabular}}
\caption{Comparison of Average Recall metric per object, on One Model per Object and One Model per Dataset, on LM-O for models trained with KeyGNet and symmetric keypoints using the total loss $\mathcal{L}_{\text{total}}$ (Eq.~\ref{eq:t_loss}).
}
\label{tab:SISO_MIMO_ar_performance}
\end{table}

Table~\ref{tab:SISO_MIMO_ar_performance} reports the Average Recall scores on LM-O for multiple objects. Notably, the Eggbox object trained with symmetric keypoints, shows a significant drop in AR under the single model setup (0.579) compared to training a separate model per object (0.691). This drop is caused by a mismatch in training objectives: the single model is trained with the total loss $\mathcal{L}_{\text{total}}$ (Eq.~\ref{eq:t_loss}), whereas symmetric objects are more effectively trained using only the radial loss $\mathcal{L}_{R}$ (Eq.~\ref{eq:l_r}).  Glue is not strictly symmetric, as the front and back labels differ and introduce asymmetry in the radiometric mode. Treating it as symmetric can therefore degrade performance, which explains the observed results.
As a result, the shared model is less effective in preserving the geometric constraints specific to symmetric object. Overall, the mean AR for the single model approach on LM-O is 0.784, which is 1.6\% lower than the 0.800 mean AR achieved when training individual models per object.

\hfill \break
\noindent\textbf{Effect of Training models with segmented objects.}
In this ablation study, we evaluate the effect of using segmentation masks versus object bounding boxes as input regions for radial map regression. The model is trained separately for objects in the LM-O dataset, using the total loss $\mathcal{L}{\text{total}}$ (Eq.~\ref{eq:t_loss}) for asymmetric objects.
Performance is assessed using the ADD-S and AR metrics.

On segmented object ROIs, the model demonstrates improved radial map regression, which in turn enhances surface representation and overall pose estimation accuracy. As shown in Table~\ref{tab:seg_vs_bbox_performance}, the ADD(-S) scores are consistently higher across the objects when using segmented regions compared to bounding box inputs. 

\begin{table}[h]
\centering
\renewcommand{\arraystretch}{1}
\resizebox{1.0\linewidth}{!}{
\begin{tabular}{lcccc}
\toprule
Object & Segmented Object ROI & Object Bounding Box ROI \\ 
\midrule
ape & \textbf{87.1} & \underline{83.9} \\
can & \textbf{96.8} & \underline{95.9} \\
cat & \textbf{71.0} & \underline{66.9} \\
driller & \textbf{97.3} & \underline{95.6} \\
\bottomrule
\end{tabular}}
\caption{Comparison of ADD(-S) scores on the LM-O dataset for models trained with KeyGNet and symmetric keypoints, using segmented object regions and object bounding boxes as input ROIs.
}
\label{tab:seg_vs_bbox_performance}
\end{table}

\begin{table}[h]
\centering
\renewcommand{\arraystretch}{1.4}
\Large
\resizebox{\linewidth}{!}{
\begin{tabular}{lcccccccc}
\toprule
\multirow{2}{*}{Object} & \multicolumn{4}{c}{Segmented Object ROI} & \multicolumn{4}{c}{Object Bounding Box ROI} \\ 
\cmidrule(lr){2-5} \cmidrule(lr){6-9}  
& $\textit{AR}_{VSD}$ & $\textit{AR}_{MSSD}$ & $\textit{AR}_{MSPD}$ & $\textit{Mean AR}$ & $\textit{AR}_{VSD}$ & $\textit{AR}_{MSSD}$ & $\textit{AR}_{MSPD}$ & $\textit{Mean AR}$ \\ 
\midrule
ape & \textbf{0.638}  & \textbf{0.830} & \textbf{0.912} & \textbf{0.795} & \underline{0.607} & \underline{0.789} &	\underline{0.891} &	\underline{0.763} \\
can & \textbf{0.737} & \textbf{0.914} & \textbf{0.921} & \textbf{0.857} & \underline{0.726} & \underline{0.895} &	\underline{0.898} & \underline{0.834} \\
cat & \textbf{0.568} & \textbf{0.713} & \textbf{0.797} & \textbf{0.693} & \underline{0.525} & \underline{0.667} & \underline{0.768} & \underline{0.654} \\
driller & \textbf{0.788} & \textbf{0.937} & \textbf{0.919} & \textbf{0.881} & \underline{0.764} & \underline{0.922} & \underline{0.901} & \underline{0.862} \\
\bottomrule
\end{tabular}}
\caption{Comparison of per-object Average Recall on the LM-O dataset for models trained with KeyGNet and symmetric keypoints, using segmented object regions and object bounding boxes as input ROIs.
}
\label{tab:seg_vs_bbox_ar_performance}
\end{table}

Similarly, Table~\ref{tab:seg_vs_bbox_ar_performance} shows that Average Recall metrics, including $\textit{AR}_{VSD}$, $\textit{AR}_{MSSD}$, and $\textit{AR}_{MSPD}$, are superior for each object when the model is trained with segmented object ROIs. These results highlight the benefit of providing the model with more precise object boundaries during training, enabling it to learn more accurate radial offsets.

\hfill \break
\noindent\textbf{Effect of Number of Keypoints in DLT Estimation.}
In this ablation study, we evaluate how varying the number of keypoints influences the Direct Linear Transform (DLT) method for object surface and pose estimation. The models are individually trained on a few objects from the LM-O dataset, incrementally using 4, 8, 12, and 16 keypoints. Each model is assessed based on its performance in predicting accurate object poses, employing the ADD-S and Average Recall metric. 

\begin{table}[t]
\centering
\renewcommand{\arraystretch}{1.4}
\resizebox{\linewidth}{!}{
\begin{tabular}{lcccccccc}
\toprule
Object & 4 Keypoints & 8 Keypoints & 12 Keypoints & 16 Keypoints \\ 
\midrule
ape & \textbf{87.1} & \underline{85.0} & 83.6 &	\underline{85.0} \\
can & \underline{96.8}	& 96.5 & \textbf{98.5} & 97.1 \\
cat & \textbf{71.0} & \underline{70.4} & 70.2 & 68.9	 \\
driller & \textbf{97.3} & 94.2 & 95.4 & \underline{95.9} \\
\bottomrule
\end{tabular}
}
\caption{ADD-S of four LM-O objects with DLTPose as the number of KeyGNet keypoints increases from 4 to 16.
}
\label{tab:multiple_kpts_adds}
\end{table}

\begin{table}[h]
\centering
\renewcommand{\arraystretch}{1.4}
\resizebox{\linewidth}{!}{
\begin{tabular}{lcccc}
\toprule
Object & 4 Keypoints & 8 Keypoints & 12 Keypoints & 16 Keypoints \\ 
\midrule
ape & \textbf{79.5} & \underline{77.6} & 76.2 & 77.1 \\
can & \underline{85.7} & 84.0 & \textbf{87.9} & 84.7 \\
cat & \textbf{69.3} & \underline{69.9} & 68.5 & 68.1 \\
driller & \underline{88.1} & 86.9 & 87.7 & \textbf{88.5} \\
\bottomrule
\end{tabular}
}
\caption{Average Recall of four LM-O objects with DLTPose as the number of KeyGNet keypoints increases from 4 to 16.
}
\label{tab:multiple_kpts_ar}
\end{table}

As shown in Tables~\ref{tab:multiple_kpts_adds} and~\ref{tab:multiple_kpts_ar}, models trained with only 4 keypoints consistently achieved either the best or second-best performance. We attribute this to reduced per-channel error in the radial regression task when using fewer keypoints. As the number of keypoints increases, the likelihood of noisy or inconsistent radial predictions across channels grows, potentially degrading overall accuracy. Therefore, the main results reported in this paper are derived using the 4-keypoint configuration.

\section{Accuracy Results Per Object}
\label{sec:ablation_results}

\begin{figure}[h]
    \centering
    \includegraphics[width=1\linewidth]{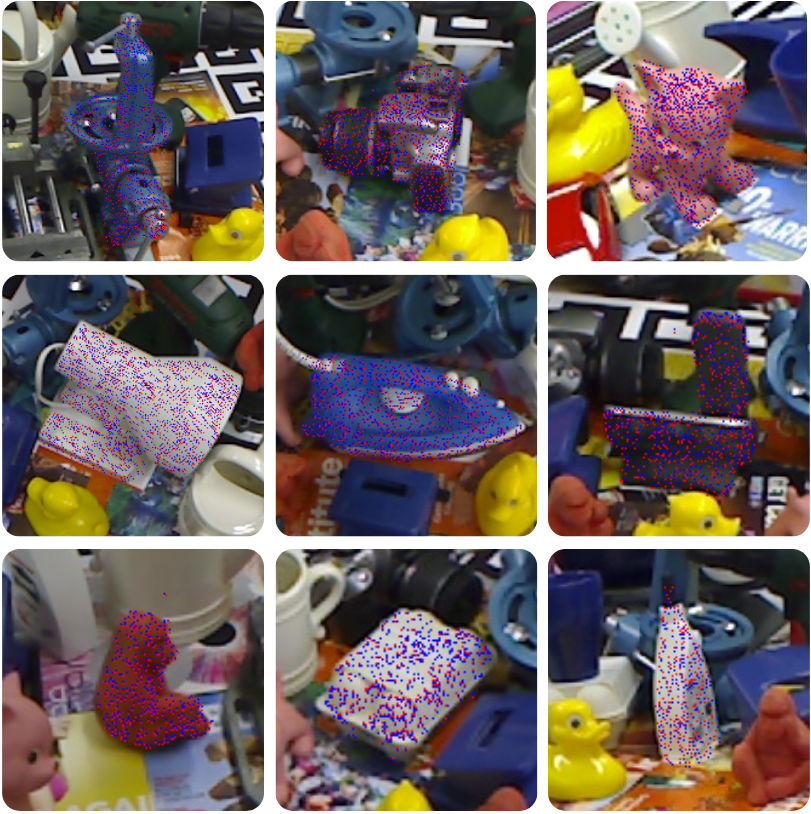}
    \caption{Overlay results on select LM images.}
    \label{fig:lm_q}

\end{figure}

\begin{figure}[h]
    \centering
    \includegraphics[width=1\linewidth]{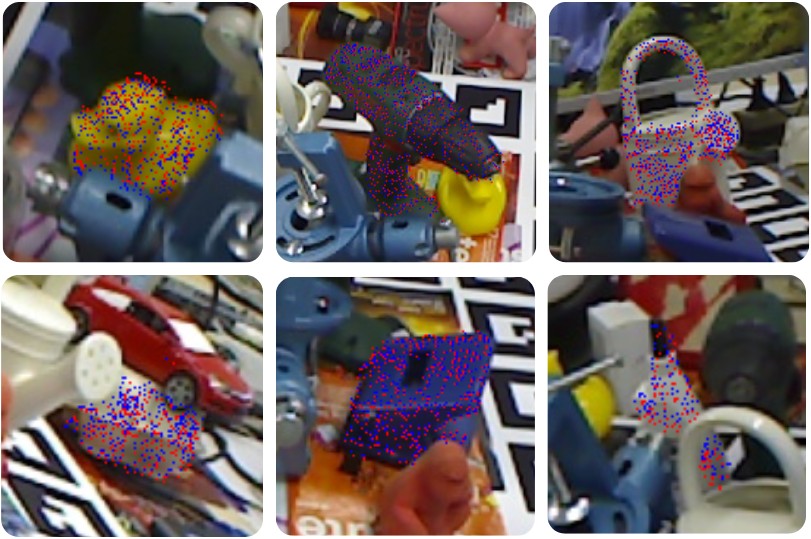}
    \caption{Overlay results on select LM-O images.}
    \label{fig:lmo_q}

\end{figure}

\begin{figure}[h]
    \centering
    \includegraphics[width=1\linewidth]{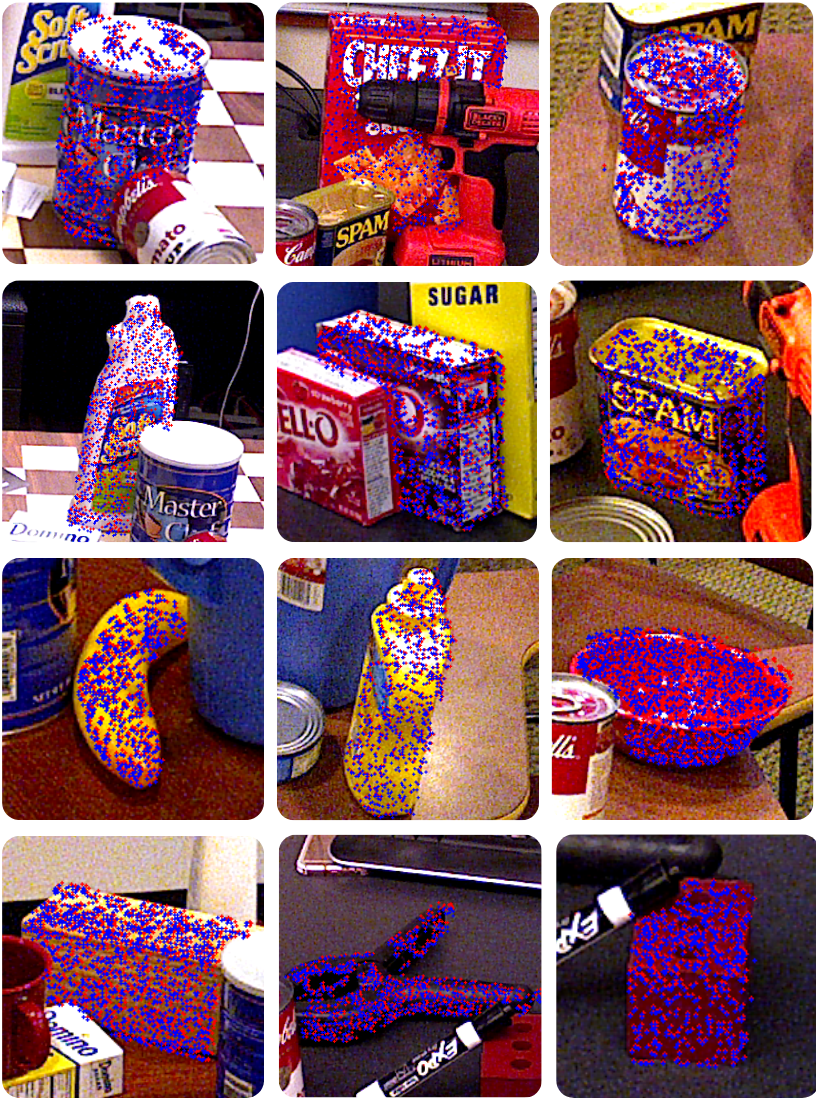}
    \caption{Overlay results on select YCB-V images.}
    \label{fig:ycbv_q}

\end{figure}

\begin{figure}[h]
        \centering
        \includegraphics[width=1\linewidth]{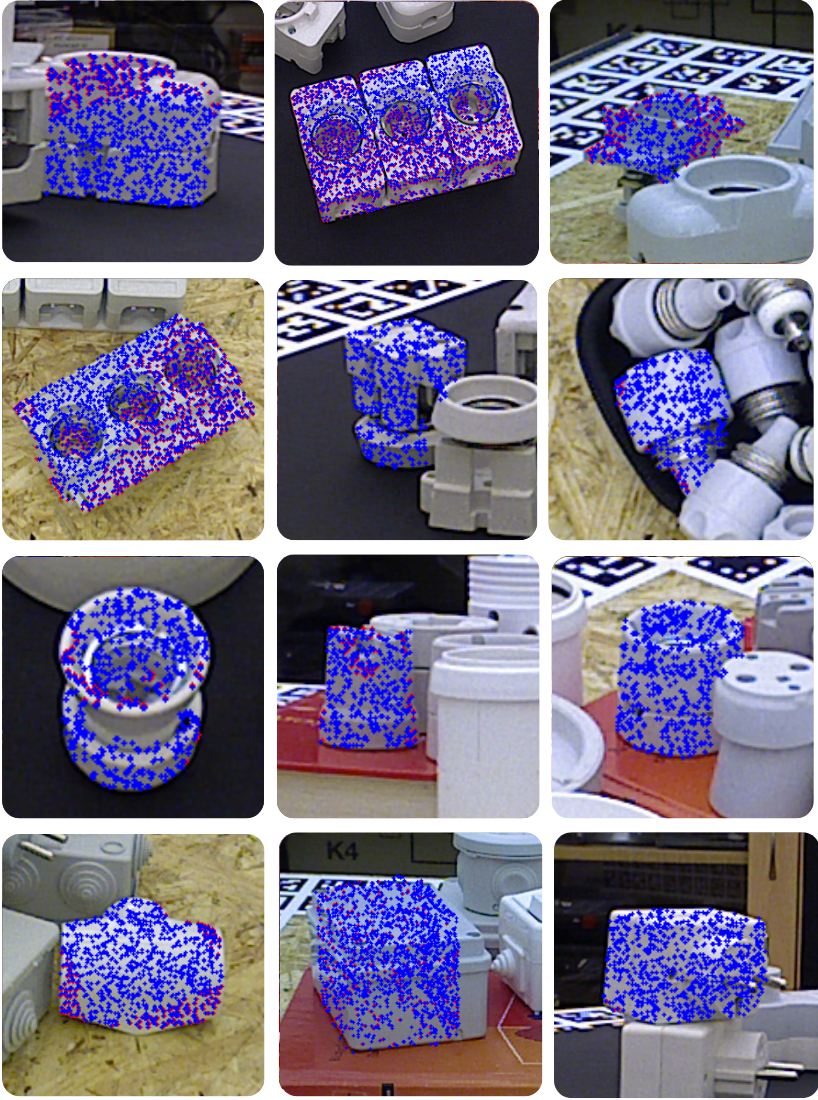}
        \caption{Overlay results on select T-LESS images.}
        \label{fig:tless_q}
\end{figure}

The detailed ADD(-S) results for the LM and  LM-O datasets, along with the AUC results for ADD-S and ADD(-S) on the YCB-V dataset, and the per object Average Recall on the T-LESS dataset are provided in Tables~\ref{tab:performance_lm}--\ref{tab:performance_tless}. Additionally, Figure~\ref{fig:lm_q}--~\ref{fig:tless_q} presents qualitative examples of successful pose recoveries, where red dots represent projected surface points from ground truth poses, while blue dots correspond to those from the estimated poses.

As shown in Table~\ref{tab:performance_lm}, the LM dataset is largely saturated, with multiple methods achieving near-perfect scores across most objects. However, our approach still achieves the highest mean ADD(-S) of 99.9\%, with perfect scores (100\%) on several objects, demonstrating its robustness in handling both symmetric and non-symmetric objects.

Table~\ref{tab:performance_lmo} highlights the challenges posed by the LM-O dataset, where occlusions and imperfect object meshes introduce significant difficulties for pose estimation. Despite these challenges, our method achieves the highest mean accuracy of 91.0\%, outperforming prior state-of-the-art approaches and showcasing strong robustness in occluded scenarios.

For the YCB-V dataset, Table~\ref{tab:performance_ycbv} reports AUC for both ADD-S and ADD(-S). Our approach achieves a mean AUC of 99.7\% for ADD-S and 97.4\% for ADD(-S), marking a significant improvement 
over prior state-of-the-art methods.

For the T-LESS dataset, Table~\ref{tab:performance_tless} presents the per-object Average Recall of our method in comparison to other approaches. T-LESS poses particular challenges due to depth-RGB misalignment and the predominance of texture-less objects, which complicates feature extraction. Among the 30 objects, our method achieved the best performance on 12 objects and ranked second on 3 objects, resulting in the highest overall mean AR of 0.839 across the dataset.

\begin{table*}[b]
\caption{Per object comparison of ADD(-S) results on LM dataset.}
\label{tab:performance_lm}
\centering

\renewcommand{\arraystretch}{1.2}
\setlength{\tabcolsep}{3pt}

\resizebox{\textwidth}{!}{%
\begin{tabular}{l
>{\centering\arraybackslash}p{1.5cm}
>{\centering\arraybackslash}p{1.5cm}
>{\centering\arraybackslash}p{1.3cm}
>{\centering\arraybackslash}p{1.0cm}
>{\centering\arraybackslash}p{1.5cm}
>{\centering\arraybackslash}p{1.3cm}
>{\centering\arraybackslash}p{1.0cm}
>{\centering\arraybackslash}p{1.0cm}
>{\centering\arraybackslash}p{1.3cm}}
\toprule
\rotatebox{45}{Object} &
\rotatebox{45}{\shortstack[l]{Dense-Fusion\\\cite{wang2019densefusion}}} &
\rotatebox{45}{\shortstack[l]{RCVPose\\\cite{Wu}}} &
\rotatebox{45}{\shortstack[l]{PVN3D\\\cite{He}}} &
\rotatebox{45}{\shortstack[l]{REDE\\\cite{Hua2020REDEEO}}} &
\rotatebox{45}{\shortstack[l]{Masked-Fusion\\\cite{pereira2020maskedfusion}}} &
\rotatebox{45}{\shortstack[l]{PR-GCN\\\cite{Zhou2021PRGCNAD}}} &
\rotatebox{45}{\shortstack[l]{EANet\\\cite{ea-net}}} &
\rotatebox{45}{\shortstack[l]{IRPE\\\cite{Jin2024IRPEIR}}} &
\rotatebox{45}{\shortstack[l]{DLTPose}} \\
\midrule
ape         & 92.3 & 99.6 & 97.3 & 95.6 & 92.2 & 99.2 & 95.1 & 95.7 & \textbf{100} \\
benchvise   & 93.2 & 99.7 & 99.7 & 99.4 & 98.4 & 99.8 & 97.5 & \underline{99.9} & \textbf{100} \\
camera      & 94.4 & \underline{99.7} & \underline{99.7} & 99.6 & 98.0 & \textbf{100} & 98.5 & 97.2 & \textbf{100} \\
can         & 93.1 & \underline{99.3} & 98.0 & 97.8 & 97.4 & \underline{99.3} & 97.7 & \textbf{100} & \textbf{100} \\
cat         & 96.5 & 99.7 & \underline{99.9} & 99.5 & 97.7 & 99.8 & 97.7 & 99.6 & \textbf{100} \\
driller     & 87.0 & \textbf{100.0} & 99.3 & 99.3 & 95.6 & \underline{99.9} & 99.2 & 99.7 & 99.8 \\
duck        & 92.3 & \underline{99.7} & 99.4 & 98.0 & 94.0 & 98.2 & 97.3 & 97.9 & \textbf{100} \\
eggbox$^*$  & 99.8 & 99.3 & 99.3 & 98.6 & 99.6 & \underline{99.9} & 99.6 & 99.7 & \textbf{100} \\
glue$^*$    & \textbf{100} & 99.7 & \textbf{100} & \textbf{100} & \textbf{100} & \textbf{100} & \underline{99.9} & 99.7 & \textbf{100} \\
holepuncher & 92.1 & \textbf{100} & 99.3 & 98.6 & 97.3 & 99.0 & 96.8 & 98.3 & \underline{99.9} \\
iron        & 97.0 & \underline{99.9} & 99.7 & 99.8 & 97.1 & \textbf{100} & 99.4 & 98.3 & 99.0 \\
lamp        & 95.3 & \underline{99.5} & \underline{99.5} & 99.3 & 99.0 & \textbf{100} & 99.2 & 98.2 & \textbf{100} \\
phone       & 92.8 & 99.7 & 99.5 & 99.5 & 98.8 & \underline{99.9} & 98.7 & 98.2 & \textbf{100} \\
\midrule
Mean        & 94.3 & \underline{99.7} & 99.4 & 98.9 & 97.3 & 99.6 & 97.6 & 98.8 & \textbf{99.9} \\
\bottomrule
\end{tabular}%
}
\end{table*}

\nolinenumbers
\begin{table*}[b]
\caption{Per object comparison of ADD or ADD-S results on LM-O. Asymmetric objects are evaluated with ADD, and symmetric objects (annotated with $\ast$) are evaluated with ADD-S.}
\label{tab:performance_lmo}
\centering

\renewcommand{\arraystretch}{1.2}
\setlength{\tabcolsep}{3pt} 

\resizebox{\textwidth}{!}{%
\begin{tabular}{l*{11}{>{\centering\arraybackslash}p{1.15cm}}}
\toprule
\rotatebox{45}{Object} &
\rotatebox{45}{\shortstack[l]{PoseCNN\\\cite{xiang2018posecnn}}} &
\rotatebox{45}{\shortstack[l]{PVN3D\\\cite{He}}} &
\rotatebox{45}{\shortstack[l]{RCVPose\\\cite{Wu}}} &
\rotatebox{45}{\shortstack[l]{REDE\\\cite{Hua2020REDEEO}}} &
\rotatebox{45}{\shortstack[l]{PR-GCN\\\cite{Zhou2021PRGCNAD}}} &
\rotatebox{45}{\shortstack[l]{ZebraPose\\\cite{su2022zebrapose}}} &
\rotatebox{45}{\shortstack[l]{IRPE\\\cite{Jin2024IRPEIR}}} &
\rotatebox{45}{\shortstack[l]{EPRO-GDR\\\cite{10896093}}} &
\rotatebox{45}{\shortstack[l]{DFTr\\\cite{zhou2023deep}}} &
\rotatebox{45}{\shortstack[l]{HiPose\\\cite{lin2024hipose}}} &
\rotatebox{45}{DLTPose} \\
\midrule
ape        & 76.2 & 33.9 & 61.3 & 53.1 & 40.2 & 60.4 & 69.8 & 64.1 & 76.6 & \underline{78.0} & \textbf{87.1} \\
can        & 87.4 & 88.6 & 93.0 & 88.5 & 76.2 & 95.0 & 95.4 & \underline{98.5} & 96.1 & \textbf{98.9} & 96.8 \\
cat        & 52.2 & 39.1 & 51.2 & 35.9 & 57.0 & 62.1 & 72.4 & \textbf{88.9} & 52.2 & \underline{87.5} & 71.0 \\
driller    & 90.3 & 78.4 & 78.8 & 77.8 & 82.3 & 94.8 & 93.0 & 96.5 & 95.8 & \textbf{97.8} & \underline{97.3} \\
duck       & 77.7 & 41.9 & 53.4 & 46.2 & 30.0 & 64.5 & 81.1 & 82.8 & 72.3 & \underline{85.3} & \textbf{88.5} \\
eggbox*    & 72.2 & 80.9 & 82.3 & 71.8 & 68.2 & 73.8 & 85.3 & 74.4 & 75.3 & 80.3 & \textbf{98.6} \\
glue*      & 76.7 & 68.1 & 72.9 & 75.0 & 67.0 & 88.7 & 88.4 & \textbf{95.0} & 79.3 & \underline{94.1} & 91.1 \\
holepuncher& 91.4 & 74.7 & 75.8 & 75.5 & 97.2 & 88.4 & \textbf{97.9} & 96.5 & 86.8 & 95.2 & \underline{97.5} \\
\midrule
Mean       & 78.0 & 63.2 & 71.1 & 65.4 & 64.8 & 78.5 & 85.4 & 88.7 & 77.7 & \underline{89.6} & \textbf{91.0} \\
\bottomrule
\end{tabular}%
}
\end{table*}
\linenumbers

\begin{sidewaystable*}[p]
\caption{Per object comparison of AUC of ADD-S/ADD(-S) results on YCB-V dataset. 
}
\label{tab:performance_ycbv}
\centering
\resizebox{\textwidth}{!}{
\begin{tabular}{l*{20}{c}}
\toprule
\multirow{2}{*}{Object} &
\multicolumn{2}{c}{PoseCNN~\cite{xiang2018posecnn}} &
\multicolumn{2}{c}{PVN3D~\cite{He}} &
\multicolumn{2}{c}{ZebraPose~\cite{su2022zebrapose}} &
\multicolumn{2}{c}{EANet~\cite{ea-net}} &
\multicolumn{2}{c}{RCVPose~\cite{Wu}} &
\multicolumn{2}{c}{DFTr~\cite{zhou2023deep}} &
\multicolumn{2}{c}{HiPose~\cite{lin2024hipose}} &
\multicolumn{2}{c}{IRPE~\cite{Jin2024IRPEIR}} &
\multicolumn{2}{c}{DLTPose} \\
\cmidrule(lr){2-3}\cmidrule(lr){4-5}\cmidrule(lr){6-7}\cmidrule(lr){8-9}%
\cmidrule(lr){10-11}\cmidrule(lr){12-13}\cmidrule(lr){14-15}\cmidrule(lr){16-17}%
\cmidrule(lr){18-19}\cmidrule(lr){20-21}
& -S & (-S) & -S & (-S) & -S & (-S) & -S & (-S) & -S & (-S) & -S & (-S) & -S & (-S) & -S & (-S) & -S & (-S) \\
\midrule
002\_master\_chef\_can & 95.8 & 69.0 & 95.2 & 79.3 & 96.3 & 80.0 & 96.6 & --  & 96.2 & -- & 97.0 & 92.3 & 96.4 & 86.2 & \textbf{100} & \textbf{100} & \textbf{100} & \underline{97.5} \\
003\_cracker\_box & 91.8 & 80.7 & 94.4 & 91.5 & 93.5 & 88.6 & 96.2 & -- & 97.9 & -- & 95.9 & 93.9 & 97.7 & 96.7 & \underline{99.0} & 96.8 & \textbf{99.8} & \textbf{99.8} \\
004\_sugar\_box & 98.2 & \underline{97.2} & 97.9 & 96.9 & 95.5 & 91.6 & 97.8 & -- & 97.9 & -- & 97.1 & 95.5 & 98.2 & 97.1 & 96.8 & 95.0 & \textbf{99.9} & \textbf{99.8} \\
005\_tomato\_soup\_can & 94.5 & 94.3 & 95.9 & 89.0 & 94.4 & 90.3 & 96.1 & -- & \underline{99.0} & -- & 95.6 & 92.6 & 97.0 & 95.1 & 98.2 & 94.1 & \textbf{100} & \textbf{99.8} \\
006\_mustard\_bottle & 98.4 & 87.0 & 98.3 & 97.9 & 96.1 & 93.0 & 96.9 & -- & 98.2 & -- & 97.6 & 96.3 & 98.4 & 96.9 & \underline{99.0} & 95.3 & \textbf{100} & \textbf{100} \\
007\_tuna\_fish\_can & 98.4 & 97.9 & 96.7 & 90.7 & 97.7 & 94.8 & 97.1 & -- & 98.6 & -- & 97.3 & 94.5 & 97.8 & 96.2 & 98.0 & \underline{98.0} & \textbf{99.3} & \textbf{98.2} \\
008\_pudding\_box & 97.9 & 96.6 & 98.2 & 97.1 & 94.2 & 84.4 & 94.8 & -- & 98.1 & -- & 97.4 & 95.7 & 98.8 & 98.1 &  \textbf{100} & \textbf{100} & \textbf{100} & \textbf{100} \\
009\_gelatin\_box & 98.8 & 96.6 & 98.8 & 98.3 & 97.8 & 88.4 & 97.1 & -- & 98.4 & -- & 97.6 & 96.3 & 98.9 & 97.8 &  99.0 & \underline{98.9} & \textbf{100} & \textbf{100} \\
010\_potted\_meat\_can & 92.8 & 83.8 & 93.8 & 87.9 & 93.8 & 84.0 & 97.2 & -- & 98.4 & -- & 95.9 & 92.1 & 93.5 & 83.4 & \textbf{86.0} & 74.4 & \textbf{99.7} & \textbf{98.1} \\
011\_banana & 96.9 & 92.6 & 98.2 & 96.0 & 92.3 & 84.3 & 97.1 & -- & 98.3 & -- & 97.1 & 95.0 & \underline{98.6} & 96.3 & \textbf{99.9} & \textbf{98.9} & \textbf{99.9} & \underline{97.1} \\
019\_pitcher\_base & 97.8 & 92.3 & 97.6 & \textbf{96.9} & 89.8 & 89.0 & 98.0 & -- & 97.2 & -- & 96.0 & 93.1 & 96.8 & \underline{93.2} & 91.2 & 87.8 & \textbf{99.5} & 87.9 \\
021\_bleach\_cleanser & 96.8 & 92.3 & 97.2 & \underline{95.9} & 89.8 & 89.0 & 98.0 & -- & \underline{99.6} & -- & 96.8 & 94.9 & 97.1 & 94.0 &  79.3 & 74.0 & \textbf{99.8} & \textbf{96.9} \\
024\_bowl$\ast$ & 78.3 & 72.6 & 92.8 & 92.8 & 85.6 & 85.6 & 97.1 & -- & 96.9 & -- & 96.9 & 96.9 & 98.0 & 98.0 &  \textbf{100} & \textbf{100} & \textbf{100} & \textbf{100} \\
025\_mug & 95.1 & 91.1 & 97.7 & 96.0 & \textbf{99.9} & \textbf{99.9} & 97.6 & -- & 98.7 & -- & 97.6 & 94.9 & 98.2 & 95.7 & \underline{99.7} & \underline{99.7} & 99.4 & 88.3 \\
035\_power\_drill & \underline{98.3} & 73.1 & 97.1 & 95.7 & 95.8 & 81.8 & 94.3 & -- & 96.4 & -- & 96.9 & 95.2 & \underline{98.3} & \underline{97.4} & 98.2 & 96.4 & \textbf{99.8} & \textbf{98.1} \\
036\_wood\_block$\ast$ & 90.5 & 79.2 & 91.1 & 91.1 & 91.1 & 79.2 & 83.6 & -- & 90.7 & -- & 96.2 & 96.2 & 97.0 & 97.0 & 74.8 & 74.8 & \textbf{100} & \textbf{100} \\
037\_scissors & 92.2 & 84.8 & 92.4 & 95.0 & 87.2 & 91.9 & 94.0 & -- & 96.4 & -- & 97.2 & 93.3 & 98.3 & \underline{96.8} & \textbf{99.7} & \textbf{99.7} & \underline{99.1} & 92.1 \\
040\_large\_marker & 97.2 & 47.3 & 98.1 & 91.6 & 97.6 & 89.7 & 94.0 & -- & 96.6 & -- & 96.9 & 92.7 & \underline{98.6} & 94.3 & 97.6 & 89.7 & \textbf{99.9} & \textbf{96.1} \\
051\_large\_clamp$\ast$ & 75.4 & 52.6 & 95.6 & 95.6 & 73.6 & 75.5 & 94.0 & -- & 96.2 & -- & \underline{96.3} & \underline{96.3} & 95.9 & 95.9 & 75.5 & 75.5 & \textbf{99.9} & \textbf{99.9} \\
052\_extra\_large\_clamp$\ast$ & 73.1 & 28.7 & 90.5 & 90.5 & 83.6 & 74.8 & 94.0 & -- & 95.1 & -- & 96.4 & 96.4 & 95.6 & 95.6 & 74.8 & 74.8 & \textbf{99.9} & \textbf{99.9} \\
061\_foam\_brick$\ast$ & 97.1 & 48.3 & 98.2 & 98.2 & 92.3 & 92.3 & 94.0 & -- & 96.6 & -- & 97.3 & 97.3 & 98.6 & 98.6 & \underline{99.7} & \underline{99.7} & \textbf{100} & \textbf{100} \\
\midrule
Mean & 93.0 & 79.3 & 96.1 & 92.3 & 92.0 & 87.5 & 94.2 & -- & 97.2 & -- & 96.7 & 94.4 & 97.5 & 95.3 & 94.9 & 90.6 & \textbf{99.7} & \textbf{97.4} \\
\bottomrule
\end{tabular}
}
\end{sidewaystable*}

\begin{table*}[t]
\caption{Per-object comparison of Average Recall on the T-LESS dataset.}
\label{tab:performance_tless}
\centering
\renewcommand{\arraystretch}{1.2}
\resizebox{0.8\linewidth}{!}{
\begin{tabular}{cccccc}
\toprule
Object & SurfEmb~\cite{Haugaard} & HiPose~\cite{lin2024hipose} & EPRO-GDR~\cite{10896093} & CIR~\cite{lipson2022coupled} & DLTPose (Ours) \\ 
\midrule
1  & 0.726 & 0.731 & \textbf{0.766} & 0.699 & \underline{0.734} \\
2  & 0.668 & \textbf{0.745} & \underline{0.741} & 0.653 & 0.651 \\
3  & \textbf{0.900} & 0.836 & \underline{0.854} & 0.848 & 0.795 \\
4  & 0.605 & 0.713 & \textbf{0.804} & 0.564 & \underline{0.751} \\
5  & 0.941 & 0.919 & \textbf{0.948} & \underline{0.946} & 0.924 \\
6  & \textbf{0.963} & 0.941 & \underline{0.946} & 0.918 & 0.939 \\
7  & 0.893 & \underline{0.909} & 0.895 & 0.628 & \textbf{0.920} \\
8  & \textbf{0.950} & \underline{0.937} & 0.900 & 0.877 & 0.915 \\
9  & \textbf{0.950} & \underline{0.944} & 0.906 & \underline{0.944} & 0.925 \\
10 & 0.934 & 0.927 & 0.925 & 0.919 & \textbf{0.952} \\
11 & \underline{0.912} & 0.869 & \textbf{0.924} & 0.866 & 0.890 \\
12 & \underline{0.921} & 0.893 & 0.910 & 0.877 & \textbf{0.944} \\
13 & 0.817 & 0.780 & \underline{0.854} & 0.723 & \textbf{0.872} \\
14 & 0.812 & \textbf{0.832} & 0.183 & \underline{0.820} & 0.806 \\
15 & 0.769 & \textbf{0.835} & 0.558 & \underline{0.804} & 0.751 \\
16 & \textbf{0.894} & \underline{0.887} & 0.448 & 0.772 & 0.831 \\
17 & \textbf{0.956} & 0.948 & 0.927 & \underline{0.955} & 0.939 \\
18 & \underline{0.918} & \textbf{0.921} & 0.914 & 0.848 & 0.814 \\
19 & 0.863 & \textbf{0.867} & 0.698 & \underline{0.865} & 0.864 \\
20 & \underline{0.815} & 0.811 & 0.653 & 0.791 & \textbf{0.850} \\
21 & 0.798 & 0.765 & \underline{0.820} & 0.728 & \textbf{0.881} \\
22 & 0.749 & 0.720 & \underline{0.758} & 0.659 & \textbf{0.837} \\
23 & \underline{0.930} & 0.915 & 0.867 & 0.909 & \textbf{0.972} \\
24 & \underline{0.948} & 0.834 & 0.905 & 0.907 & \textbf{0.959} \\
25 & 0.831 & \underline{0.895} & \textbf{0.924} & 0.862 & 0.835 \\
26 & \underline{0.937} & 0.869 & \textbf{0.947} & 0.862 & 0.912 \\
27 & \underline{0.906} & 0.891 & 0.125 & 0.811 & \textbf{0.915} \\
28 & \underline{0.929} & 0.926 & 0.417 & 0.877 & \textbf{0.936} \\
29 & \underline{0.948} & 0.908 & 0.929 & 0.585 & \textbf{0.949} \\
30 & 0.794 & \textbf{0.885} & 0.195 & 0.658 & \underline{0.832} \\
\bottomrule
\end{tabular}}
\end{table*}

\end{document}